\documentclass[a4paper,twoside]{article}

\usepackage{epsfig}
\usepackage{subcaption}
\usepackage{calc}
\usepackage{amssymb}
\usepackage{amstext}
\usepackage{amsmath}
\usepackage{amsthm}
\usepackage{multicol}
\usepackage{pslatex}
\usepackage{apalike}
\usepackage{algorithm2e}
\usepackage[bottom]{footmisc}

\usepackage{graphicx}
\usepackage{booktabs}
\usepackage[nolist]{acronym}
\usepackage{multirow}
\usepackage{xcolor}
\usepackage{siunitx}
\usepackage{todonotes}
\usepackage{float}

\usepackage{amsmath,amsfonts,bm}
\usepackage{mathtools}

\def\eqref#1{equation~\ref{#1}}

\def\1{\bm{1}}

\def\rvx{{\mathbf{x}}}

\DeclareMathAlphabet{\mathsfit}{\encodingdefault}{\sfdefault}{m}{sl}
\SetMathAlphabet{\mathsfit}{bold}{\encodingdefault}{\sfdefault}{bx}{n}

\newcommand{\E}{\mathbb{E}}

\def\imagescale{1.1}

\renewcommand{\paragraph}[1]{{\vskip 6pt \noindent\textbf{#1.} }}
\newcommand\blfootnote[1]{%
  \begingroup
  \renewcommand\thefootnote{}\footnote{#1}%
  \addtocounter{footnote}{-1}%
  \endgroup
}

\usepackage[hidelinks]{hyperref}
\PassOptionsToPackage{hyphens}{url}\usepackage{hyperref}

\usepackage[capitalize]{cleveref}
\crefname{section}{Sec.}{Secs.}
\Crefname{section}{Section}{Sections}
\Crefname{table}{Table}{Tables}
\crefname{table}{Tab.}{Tabs.}

\usepackage{SCITEPRESS_arxiv} %

\begin{document}

\begin{acronym}
    \acro{GAN}{generative adversarial network}
    \acro{DM}{diffusion model}
    \acro{DFT}{discrete Fourier transform}
    \acro{DCT}{discrete cosine transform}
    \acro{CNN}{convolutional neural network}
    \acro{FID}{Fr\'echet inception distance}
    \acro{AUROC}{area under the receiver operating characteristic curve}
    \acro{Pd@FAR}{probability of detection at a fixed false alarm rate}
    \acro{MSE}{mean squared error}
\end{acronym}

\title{Towards the Detection of Diffusion Model Deepfakes}

\author{\authorname{Jonas Ricker\sup{1}\orcidAuthor{0000-0002-7186-3634}, Simon Damm\sup{1}\orcidAuthor{0000-0002-4584-1765}, Thorsten Holz\sup{2}\orcidAuthor{0000-0002-2783-1264} and Asja Fischer\sup{1}\orcidAuthor{0000-0002-1916-7033}}
\affiliation{\sup{1}Ruhr University Bochum, Bochum, Germany}
\affiliation{\sup{2}CISPA Helmholtz Center for Information Security, Saarbrücken, Germany}
\email{\{jonas.ricker, simon.damm, asja.fischer\}@rub.de, holz@cispa.de}
}


\abstract{In the course of the past few years, diffusion models (DMs) have reached an unprecedented level of visual quality. However, relatively little attention has been paid to the detection of DM-generated images, which is critical to prevent adverse impacts on our society. In contrast, generative adversarial networks (GANs), have been extensively studied from a forensic perspective. In this work, we therefore take the natural next step to evaluate whether previous methods can be used to detect images generated by DMs. Our experiments yield two key findings: (1) state-of-the-art GAN detectors are unable to reliably distinguish real from DM-generated images, but (2) re-training them on DM-generated images allows for almost perfect detection, which remarkably even generalizes to GANs. Together with a feature space analysis, our results lead to the hypothesis that DMs produce fewer detectable artifacts and are thus more difficult to detect compared to GANs. One possible reason for this is the absence of grid-like frequency artifacts in DM-generated images, which are a known weakness of GANs. However, we make the interesting observation that diffusion models tend to underestimate high frequencies, which we attribute to the learning objective.}

\onecolumn \maketitle \normalsize \setcounter{footnote}{0} \vfill

\section{\uppercase{Introduction}}
\blfootnote{This is the extended version of the paper that was accepted at VISAPP 2024.}
\label{sec:intro}
In the recent past, \acfp{DM} have shown a lot of promise as a method for synthesizing images. Such models provide better (or at least similar) performance compared to \acfp{GAN} and enable powerful text-to-image models such as DALL\textperiodcentered E 2~\cite{rameshHierarchicalTextconditionalImage2022}, Imagen~\cite{sahariaPhotorealisticTexttoimageDiffusion2022}, and Stable Diffusion~\cite{rombachHighresolutionImageSynthesis2022}. 
Advances in image synthesis have resulted in very high-quality images being generated, and humans can hardly tell if a given picture is an actual or artificially generated image (so-called \emph{deepfake})~\cite{nightingaleAIsynthesizedFacesAre2022}.
This progress has many implications in practice and poses a danger to our digital society: Deepfakes can be used for disinformation campaigns, as such images appear particularly credible due to their sensory comprehensibility.
Disinformation aims to discredit opponents in public perception, to create sentiment for or against certain social groups, and thus influence public opinion. 
In effect, deepfakes lead to an erosion of trust in institutions or individuals, support conspiracy theories, and promote a fundamental political camp formation.
\ac{DM}-based text-to-image models entail particular risks, since an adversary can specifically create images supporting their narrative, with very little technical knowledge required.
A recent example of public deception featuring \ac{DM}-generated images---although without malicious intent---is the depiction of Pope Francis in a puffer jacket~\cite{huangWhyPopeFrancis2023}.
Despite the growing concern about deepfakes and the continuous improvement of \acp{DM}, there is only a limited amount of research on their detection.

In this paper, we conduct an extensive experimental study on the detectability of images generated by \acp{DM}.
Since previous work on the detection of \ac{GAN}-generated images (e.g.,~\cite{wangCNNgeneratedImagesAre2020,gragnanielloAreGANGenerated2021,mandelliDetectingGANgeneratedImages2022}) resulted in effective detection methods, we raise the question whether these can be applied to \ac{DM}-generated images.
Our analysis on five state-of-the-art \acp{GAN} and five \acp{DM} demonstrates that existing detection methods suffer from severe performance degradation when applied to \ac{DM}-generated images, with the \acs{AUROC} dropping by \SI{15.2}{\percent} on average compared to \acp{GAN}.
However, we show that by re-training, the detection accuracy can be drastically improved, proving that images generated by \acp{DM} \textit{can} be detected.
Remarkably, a detector trained on \ac{DM}-generated images is capable of detecting images from \acp{GAN}, while the opposite direction does not hold.
Our analysis in feature space suggests that \ac{DM}-generated images are harder to detect because they contain fewer generation artifacts, particularly in the frequency domain.
However, we observe a previously overlooked mismatch towards higher frequencies.
Further analysis suggests that this is caused by the training objective of \acp{DM}, which favors perceptual image quality instead of accurate reproduction of high-frequency details.
We believe that our results provide the foundation for further research on the effective detection of deepfakes generated by \acp{DM}.
Our code and data are available at \url{https://github.com/jonasricker/diffusion-model-deepfake-detection}.

\section{\uppercase{Related Work}}
\label{sec:related}

\paragraph{Fake Image Detection}
In the wake of the emergence of powerful image synthesis methods, the forensic analysis of deepfake images received increased attention, leading to a variety of detection methods~\cite{verdolivaMediaForensicsDeepFakes2020}.
Existing approaches can be broadly categorized into two groups.
Methods in the first group exploit either semantic inconsistencies like irregular eye reflections~\cite{huExposingGANGeneratedFaces2021} or known generation artifacts in the spatial~\cite{natarajDetectingGANGenerated2019,mccloskeyDetectingGANgeneratedImagery2019} or frequency domain~\cite{frankLeveragingFrequencyAnalysis2020}.
The second group uses neural networks to learn a feature representation in which real images can be distinguished from generated ones.
Wang et al.\ demonstrate that training a standard \ac{CNN} on real and fake images from a single \ac{GAN} yields a classifier capable of detecting images generated by a variety of unknown \acp{GAN} \cite{wangCNNgeneratedImagesAre2020}.
Given the rapid evolution of generative models, developing detectors which generalize to new generators is crucial and therefore a major field of research~\cite{xuanGeneralizationGANImage2019,chaiWhatMakesFake2020,wangCNNgeneratedImagesAre2020,cozzolinoUniversalGANImage2021,gragnanielloAreGANGenerated2021,girishDiscoveryAttributionOpenworld2021,mandelliDetectingGANgeneratedImages2022,jeongFingerprintNet2022}.

Since \acp{DM} have been proposed only recently, few works analyze their forensic properties.
Farid performs an initial exploration of lighting~\cite{faridLightingConsistencyPaint2022} and perspective~\cite{faridPerspectiveConsistencyPaint2022} inconsistencies in images generated by DALL\textperiodcentered E 2~\cite{rameshHierarchicalTextconditionalImage2022}, showing that \acp{DM} often generate physically implausible scenes.
A novel approach specifically targeted at \acp{DM} is proposed in \cite{wangDIREDiffusiongeneratedImage2023}, who observe that \ac{DM}-generated images can be more accurately reconstructed by a pre-trained \ac{DM} than real images.
The difference between the original and reconstructed image then serves as the input for a binary classifier.
Another work~\cite{shaDEFAKEDetectionAttribution2022} focuses on text-to-image models like Stable Diffusion~\cite{rombachHighresolutionImageSynthesis2022}.
They find that incorporating the prompt with which an image was generated (or a generated caption if the real prompt is not available) into the detector improves classification.
In a work related to ours \cite{corviDetectionSyntheticImages2023}, it is shown that \ac{GAN} detectors perform poorly on \ac{DM}-generated images.
Therefore, a pressing challenge is to develop \textit{universal} detection methods that are effective against different kinds of generative models, mainly \acp{GAN} and \acp{DM}.
Ojha et al.\ make a first step in this direction \cite{ojhaUniversalFakeImage2023}.
Instead of training a classifier directly on real and fake images, which according to their hypothesis leads to poor generalization since the detector focuses on e.g., \ac{GAN}-specific artifacts, they propose to use a pre-trained vision transformer (CLIP-ViT~\cite{dosovitskiyImageWorth16x162021,radfordLearningTransferableVisual2021}), extended with a final classification layer.

\paragraph{Frequency Artifacts in Generated Images}
Zhang et al.\ were the first to demonstrate that the spectrum of \ac{GAN}-generated images contains visible artifacts in the form of a periodic, grid-like pattern due to transposed convolution operations \cite{zhangDetectingSimulatingArtifacts2019}.
These findings were later reproduced~\cite{wangCNNgeneratedImagesAre2020} and extended to the~\ac{DCT}~\cite{frankLeveragingFrequencyAnalysis2020}.
Another characteristic was discovered in \cite{durallWatchYourUpconvolution2020}, who showed that \acp{GAN} are unable to correctly reproduce the spectral distribution of the training data.
In particular, generated images contain increased magnitudes at high frequencies.
While several works attribute these spectral discrepancies to transposed convolutions~\cite{zhangDetectingSimulatingArtifacts2019,durallWatchYourUpconvolution2020} or, more general, up-sampling operations~\cite{frankLeveragingFrequencyAnalysis2020,chandrasegaranCloserLookFourier2021}, no consensus on their origin has yet been reached.
Some works explain them by the spectral bias of convolution layers due to linear dependencies~\cite{dzanicFourierSpectrumDiscrepancies2020,khayatkhoeiSpatialFrequencyBias2022}, while others suggest the discriminator is not able to provide an accurate training signal~\cite{chenSSDGANMeasuringRealness2021,schwarzFrequencyBiasGenerative2021}.

In contrast, whether images generated by \acp{DM} exhibit grid-like frequency patterns appears to strongly depend on the specific model~\cite{shaDEFAKEDetectionAttribution2022,corviIntriguingPropertiesSynthetic2023,ojhaUniversalFakeImage2023}.
Another interesting observation is made by Rissanen et al.\ who analyze the generative process of diffusion models in the frequency domain \cite{rissanenGenerativeModellingInverse2022}.
They state that diffusion models have an inductive bias according to which, during the reverse process, higher frequencies are added to existing lower frequencies.
Other works~\cite{kingmaVariationalDiffusionModels2021,songScorebasedGenerativeModeling2022} experiment with adding Fourier features to improve learning of high-frequency content, the former reporting it leads to much better likelihoods.

\section{\uppercase{Background on DM}s}

\label{sec:background}
\Acp{DM} are a class of probabilistic generative models, originally inspired by nonequilibrium thermodynamics \cite{sohl-dicksteinDeepUnsupervisedLearning2015}.
The most common formulations build either on DDPM \cite{hoDenoisingDiffusionProbabilistic2020} or the score-based modeling perspective \cite{songGenerativeModelingEstimating2019,songImprovedTechniquesTraining2020,songScorebasedGenerativeModeling2022}. 
Numerous modifications and improvements have been proposed, leading to higher perceptual quality~\cite{nicholImprovedDenoisingDiffusion2021,dhariwalDiffusionModelsBeat2021,choiPerceptionPrioritizedTraining2022,rombachHighresolutionImageSynthesis2022} and increased sampling speed~\cite{songDenoisingDiffusionImplicit2022,liuPseudoNumericalMethods2022,salimansProgressiveDistillationFast2022,xiaoTacklingGenerativeLearning2022}.
In short, \Acp{DM} model a data distribution by gradually disturbing a sample from this distribution and then learning to reverse this diffusion process. To be more precise, we briefly review the forward and backward process for the seminal work in \cite{hoDenoisingDiffusionProbabilistic2020}.
In the diffusion (or forward) process for DDPMs, a sample $\rvx_0$ (an image in most applications) is repeatedly corrupted by Gaussian noise in sequential steps $t=1,\dots, T$ in dependence of a monotonically increasing noise schedule $\{\beta_t\}_{t=1}^T$:
\begin{equation}
    q(\rvx_t \vert \rvx_{t-1}) = \mathcal{N}(\sqrt{1-\beta_t} \rvx_{t-1}, \beta_t \mathbf{I}) \enspace .
\end{equation}
With $\alpha_t \coloneqq 1-\beta_t$ and ${\bar \alpha_t \coloneqq \prod_{s=1}^t  \alpha_t}$, we can directly sample from the forward process at arbitrary times:
\begin{equation}
    q(\rvx_t \vert \rvx_0) = \mathcal{N}(\sqrt{\bar \alpha_t} \rvx_0, (1-\bar\alpha_t) \mathbf{I})\enspace.
\end{equation}
The noise schedule is typically designed to satisfy $q(\rvx_T \vert \rvx_0) \approx \mathcal{N}(\mathbf{0}, \mathbf{I})$.
During the denoising (or reverse) process, we aim to iteratively sample from $q(\rvx_{t-1} \vert \rvx_t)$ to ultimately obtain a clean image from $\rvx_T \sim \mathcal{N}(\mathbf{0}, \mathbf{I})$.
However, since $q(\rvx_{t-1} \vert \rvx_t)$ is intractable as it depends on the entire underlying data distribution, it is approximated by a deep neural network.
More formally, $q(\rvx_{t-1} \vert \rvx_t)$ is approximated by
\begin{equation} 
    p_\theta(\rvx_{t-1} \vert \rvx_{t}) = \mathcal{N}(\mu_\theta(\rvx_t,t), \Sigma_\theta(\rvx_t,t)) \enspace,
\end{equation}
where mean $\mu_\theta$ and covariance $\Sigma_\theta$ are given by the output of the model (or the latter is set to a constant as proposed in~\cite{hoDenoisingDiffusionProbabilistic2020}).
Predicting the mean of the denoised sample $\mu_\theta(\rvx_t,t)$ is conceptually equivalent to predicting the noise that should be removed, denoted by $\epsilon_\theta(\rvx_t,t)$.
Predominantly, the latter approach is implemented (e.g.,~\cite{hoDenoisingDiffusionProbabilistic2020,dhariwalDiffusionModelsBeat2021}) such that training a \ac{DM} boils down to minimizing a (weighted) \ac{MSE} ${\Vert \epsilon - \epsilon_\theta(\rvx_t, t) \Vert^2}$ between the true and predicted noise. 
Note that this objective can be interpreted as a weighted ELBO with data augmentation \cite{kingma2023understanding}.
For a recent overview on \acp{DM} see \cite{yang2023DiffusionReview}.

\section{\uppercase{Dataset}}
\label{sec:dataset}
To ensure technical correctness, we decide to analyze a set of generative models for which pre-trained checkpoints and/or samples of \textit{the same} dataset, namely LSUN Bedroom~\cite{yuLSUNConstructionLargescale2016} (256$\times$256), are available.
Otherwise, both the detectability of generated samples and their spectral properties might suffer from biases, making them difficult to compare.
An overview of the dataset is given in Table~\ref{tab:dataset} and we provide details and example images in \cref{app:dataset}.

\begin{table}[b]
    \centering
    \caption{{Models evaluated in this work.} \acp{FID} on LSUN Bedroom
    are taken from the original publications and from~\protect\cite{dhariwalDiffusionModelsBeat2021} in the case of IDDPM. The lower the \ac{FID}, the higher the image quality.}
    \label{tab:dataset}
    \begin{tabular}{llr}
        \toprule
        Model Class          & Method                                                                   & FID  \\ \midrule
        \multirow{5}{*}{GAN} & ProGAN                         & 8.34 \\
                             & StyleGAN               & 2.65 \\
                             & ProjectedGAN                      & 1.52 \\
                             & Diff-StyleGAN2                   & 3.65 \\
                             & Diff-ProjectedGAN                & 1.43 \\ \midrule
        \multirow{5}{*}{DM}  & DDPM                        & 6.36 \\
                             & IDDPM                       & 4.24 \\
                             & ADM                            & 1.90 \\          
                             & PNDM                               & 5.68 \\
                             & LDM                       & 2.95 \\ \bottomrule
    \end{tabular}
\end{table}

All samples are either directly downloaded or generated using code and pre-trained models provided by the original publications.
We consider data from ten models in total, five \acp{GAN} and five \acp{DM}.
This includes the seminal models ProGAN~\cite{karrasProgressiveGrowingGANs2018} and StyleGAN~\cite{karrasStylebasedGeneratorArchitecture2019}, as well as the more recent ProjectedGAN~\cite{sauerProjectedGANsConverge2021}.
Note that Diff(usion)-StyleGAN2 and Diff(usion)-ProjectedGAN~\cite{wangDiffusionGANTrainingGANs2022} (the current state of the art on LSUN Bedroom) use a forward diffusion process to optimize \ac{GAN} training, but this does not change the \ac{GAN} model architecture.
From the class of \acp{DM}, we consider the original DDPM~\cite{hoDenoisingDiffusionProbabilistic2020}, its successor IDDPM~\cite{nicholImprovedDenoisingDiffusion2021}, and ADM~\cite{dhariwalDiffusionModelsBeat2021}, the latter outperforming several \acp{GAN} with an \ac{FID}~\cite{heuselGANsTrainedTwo2017} of 1.90 on LSUN Bedroom.
PNDM~\cite{liuPseudoNumericalMethods2022} speeds up the sampling process by a factor of 20 using pseudo numerical methods, which can be applied to existing pre-trained \acp{DM}.
Lastly, LDM~\cite{rombachHighresolutionImageSynthesis2022} uses an adversarially trained autoencoder that transforms an image from the pixel space to a latent space (and back).
Training the \ac{DM} in this more suitable latent space reduces the computational complexity and therefore enables training on higher resolutions.
The success of this approach is underpinned by the groundbreaking results of Stable Diffusion, a powerful and publicly available text-to-image model based on LDM.

\begin{table*}[ht!]
    \centering
    \caption{{Detection performance of pre-trained universal detectors.} For Wang2020 and Gragnaniello2021, we consider two different variants, respectively.
    In the upper half, we report the performance of models trained on LSUN Bedroom, while results on additional datasets are given in the second half.
    The best score (determined by the highest Pd@1\%) for each generator is highlighted in {bold}. We report average scores in {\color{gray}gray}.}
    \label{tab:pretrained}
    \begin{tabular}{@{}l@{\ \ }ccccc@{}}
\toprule
 \multirow{2.75}{*}{AUROC / Pd@1\%} & \multicolumn{2}{c}{Wang2020} & \multicolumn{2}{c}{Gragnaniello2021} & Mandelli2022 \\
 \cmidrule(lr){2-3} \cmidrule(lr){4-5}
 & Blur+JPEG (0.5) & Blur+JPEG (0.1) & ProGAN & StyleGAN2 &  \\
\midrule
ProGAN & \textbf{100.0 / 100.0} & \textbf{100.0 / 100.0} & \textbf{100.0 / 100.0} & \textbf{100.0 / 100.0} & \phantom{0}91.2 / \phantom{0}27.5 \\
StyleGAN & \phantom{0}98.7 / \phantom{0}81.4 & \phantom{0}99.0 / \phantom{0}84.4 & \textbf{100.0 / 100.0} & \textbf{100.0 / 100.0} & \phantom{0}89.6 / \phantom{0}14.7 \\
ProjectedGAN & \phantom{0}94.8  / \phantom{0}49.1 & \phantom{0}90.9 / \phantom{0}34.5 & \textbf{100.0 / \phantom{0}99.3} & \phantom{0}99.9 / \phantom{0}97.8 & \phantom{0}59.4 / \phantom{0}\phantom{0}2.4 \\
Diff-StyleGAN2 & \phantom{0}99.9 / \phantom{0}97.9 & 100.0 / \phantom{0}99.3 & \textbf{100.0 / 100.0} & \textbf{100.0 / 100.0} & 100.0 / \phantom{0}99.9 \\
Diff-ProjectedGAN & \phantom{0}93.8 / \phantom{0}43.3 & \phantom{0}88.8 / \phantom{0}27.2 & \textbf{\phantom{0}99.9 / \phantom{0}99.2} & \phantom{0}99.8 / \phantom{0}96.6 & \phantom{0}62.1 / \phantom{0}\phantom{0}2.8 \\ 
\color{gray} Average & \color{gray} \phantom{0}97.4 / \phantom{0}74.3 & \color{gray} \phantom{0}95.7 / \phantom{0}69.1 & \color{gray} \textbf{100.0 / \phantom{0}99.7} & \color{gray} \phantom{0}99.9 / \phantom{0}98.9 & \color{gray} \phantom{0}80.4 / \phantom{0}29.5 \\ \midrule
DDPM & \phantom{0}85.2 / \phantom{0}14.2 & \phantom{0}80.8 / \phantom{0}\phantom{0}9.3 & \textbf{\phantom{0}96.5 /  \phantom{0}39.1} & \phantom{0}95.1 / \phantom{0}30.7 & \phantom{0}57.4 / \phantom{0}\phantom{0}0.6 \\
IDDPM & \phantom{0}81.6 /  \phantom{0}10.6 & \phantom{0}79.9 / \phantom{0}\phantom{0}7.8 & \textbf{\phantom{0}94.3 / \phantom{0}25.7} & \phantom{0}92.8 / \phantom{0}21.2 & \phantom{0}62.9 /\phantom{0}\phantom{0}1.3 \\
ADM & \phantom{0}68.3 / \phantom{0}\phantom{0}3.4 & \phantom{0}68.8 / \phantom{0}\phantom{0}4.0 & \textbf{\phantom{0}77.8 / \phantom{0}\phantom{0}5.2} & \phantom{0}70.6  / \phantom{0}\phantom{0}2.5 & \phantom{0}60.5 / \phantom{0}\phantom{0}1.8 \\
PNDM & \phantom{0}79.0 / \phantom{0}\phantom{0}9.2 & \phantom{0}75.5 / \phantom{0}\phantom{0}6.3 & \phantom{0}91.6 /  \phantom{0}16.6 & \textbf{\phantom{0}91.5 / \phantom{0}22.2} & \phantom{0}71.6 / \phantom{0}\phantom{0}4.0 \\
LDM & \phantom{0}78.7 / \phantom{0}\phantom{0}7.4 & \phantom{0}77.7 / \phantom{0}\phantom{0}6.9 & \phantom{0}96.7 / \phantom{0}42.1 & \textbf{\phantom{0}97.0 / \phantom{0}48.9} & \phantom{0}54.8 / \phantom{0}\phantom{0}2.1 \\
\color{gray} Average & \color{gray} \phantom{0}78.6 / \phantom{0}\phantom{0}9.0 & \color{gray} \phantom{0}76.6 / \phantom{0}\phantom{0}6.8 & \color{gray} \textbf{\phantom{0}91.4 / \phantom{0}25.7} & \color{gray} \phantom{0}89.4 /  \phantom{0}25.1 & \color{gray} \phantom{0}61.4 / \phantom{0}\phantom{0}2.0 \\
\midrule
\midrule
ADM (LSUN Cat) & \phantom{0}58.4 / \phantom{0}\phantom{0}2.5 & \phantom{0}58.1 / \phantom{0}\phantom{0}3.3 & \textbf{\phantom{0}60.2 / \phantom{0}\phantom{0}4.2} & \phantom{0}51.7  / \phantom{0}\phantom{0}1.8 & \phantom{0}55.6 / \phantom{0}\phantom{0}1.3 \\
ADM (LSUN Horse) & \phantom{0}55.5 / \phantom{0}\phantom{0}1.5 & \phantom{0}53.4 / \phantom{0}\phantom{0}2.2 & \textbf{\phantom{0}56.1 / \phantom{0}\phantom{0}2.7} & \phantom{0}50.2 / \phantom{0}\phantom{0}1.4 & \phantom{0}44.2 / \phantom{0}\phantom{0}0.5 \\
ADM (ImageNet) & \phantom{0}69.1 / \phantom{0}\phantom{0}4.1 & \phantom{0}71.7 / \phantom{0}\phantom{0}4.5 & \phantom{0}72.1 / \phantom{0}\phantom{0}3.5 & \textbf{\phantom{0}83.9 / \phantom{0}16.6} & \phantom{0}60.1 / \phantom{0}\phantom{0}0.9 \\
ADM-G-U (ImageNet) & \phantom{0}67.2 / \phantom{0}\phantom{0}3.7 & \phantom{0}62.3 / \phantom{0}\phantom{0}1.2 & \phantom{0}66.8 / \phantom{0}\phantom{0}1.6 & \textbf{\phantom{0}78.9 / \phantom{0}10.2} & \phantom{0}60.0 / \phantom{0}\phantom{0}1.0 \\
PNDM (LSUN Church) & \phantom{0}76.9 / \phantom{0}10.2 & \phantom{0}77.6 / \phantom{0}12.0 & \phantom{0}90.9 / \phantom{0}24.5 & \textbf{\phantom{0}99.3 / \phantom{0}85.8} & \phantom{0}56.4 / \phantom{0}\phantom{0}1.9 \\
LDM (LSUN Church) & \phantom{0}86.3 / \phantom{0}19.8 & \phantom{0}82.2 / \phantom{0}14.2 & \phantom{0}98.8 / \phantom{0}75.5 & \textbf{\phantom{0}99.5 / \phantom{0}90.2} & \phantom{0}58.9 / \phantom{0}\phantom{0}1.3 \\
LDM (FFHQ) & \phantom{0}69.4 / \phantom{0}\phantom{0}3.6 & \phantom{0}71.0 / \phantom{0}\phantom{0}3.6 & \textbf{\phantom{0}91.1 / \phantom{0}25.4} & \phantom{0}67.2 / \phantom{0}\phantom{0}2.1 & \phantom{0}63.0 / \phantom{0}\phantom{0}0.6 \\
{ADM' (FFHQ)} & \phantom{0}77.7 / \phantom{0}\phantom{0}8.7 & \phantom{0}81.4 / \phantom{0}\phantom{0}8.8 & \textbf{\phantom{0}87.7 /  \phantom{0}17.8} & \phantom{0}89.0 / \phantom{0}17.2 & \phantom{0}69.8 / \phantom{0}\phantom{0}2.0 \\
{P2 (FFHQ)} & \phantom{0}79.5 / \phantom{0}\phantom{0}8.9 & \phantom{0}83.2 / \phantom{0}\phantom{0}9.2 & \phantom{0}89.2 / \phantom{0}11.5 & \textbf{\phantom{0}91.1 / \phantom{0}18.9} & \phantom{0}72.5 / \phantom{0}\phantom{0}2.7 \\
\midrule
Stable Diffusion v1-1 & \phantom{0}42.4 / \phantom{0}\phantom{0}1.5 & \phantom{0}51.4  / \phantom{0}\phantom{0}2.0 & \phantom{0}73.2 / \phantom{0}\phantom{0}4.0 & \textbf{\phantom{0}75.2 / \phantom{0}13.6} & \phantom{0}76.1 / \phantom{0}\phantom{0}4.2 \\
Stable Diffusion v1-5 & \phantom{0}43.7 / \phantom{0}\phantom{0}1.4 & \phantom{0}52.6  / \phantom{0}\phantom{0}2.1 & \phantom{0}72.9 / \phantom{0}\phantom{0}2.8 & \textbf{\phantom{0}79.8 / \phantom{0}18.3} & \phantom{0}75.3 / \phantom{0}\phantom{0}4.1 \\
Stable Diffusion v2-1 & \textbf{\phantom{0}46.1 / \phantom{0}\phantom{0}1.4} & \phantom{0}47.3 / \phantom{0}\phantom{0}1.1 & \phantom{0}62.8 / \phantom{0}\phantom{0}1.1 & \phantom{0}55.1 / \phantom{0}\phantom{0}1.1 & \phantom{0}37.0 / \phantom{0}\phantom{0}0.5 \\
Midjourney v5 & \phantom{0}52.7 / \phantom{0}\phantom{0}3.0 & \phantom{0}57.1 / \phantom{0}\phantom{0}3.0 & \textbf{\phantom{0}69.9 / \phantom{0}\phantom{0}3.3} & \phantom{0}67.1 / \phantom{0}\phantom{0}3.3 & \phantom{0}18.3 / \phantom{0}\phantom{0}0.3 \\
\bottomrule
\end{tabular}

\end{table*}

\section{\uppercase{Detection Analysis}}
\label{sec:detection}
In this section we analyze how well state-of-the-art fake image detectors can distinguish \ac{DM}-generated from real images.
At first, we apply pre-trained detectors known to be effective against \acp{GAN}, followed by a study on the generalization abilities of re-trained detectors.
Based on our findings, we conduct an in-depth feature space analysis to gain a better understanding on how fake images are detected.

\paragraph{Detection Methods}
We evaluate three state-of-the-art \ac{CNN}-based detectors: Wang2020~\cite{wangCNNgeneratedImagesAre2020}, Gragnaniello2021~\cite{gragnanielloAreGANGenerated2021}, and Mandelli2022~\cite{mandelliDetectingGANgeneratedImages2022}.
They are supposed to perform well on images from unseen generative models, but it is unclear whether this holds for \ac{DM}-generated images as well.
In \cref{app:detectors} we provide descriptions of all three detectors.

\paragraph{Performance Metrics}
The performance of the analyzed classifiers is estimated in terms of the widely used \acf{AUROC}.
However, the \ac{AUROC} is overly optimistic as it captures merely the potential of a classifier, but the optimal threshold is usually unknown \cite{cozzolinoUniversalGANImage2021}.
Thus, we adopt the use of the \ac{Pd@FAR} as an additional metric, which is given as the true positive rate at a fixed false alarm rate.
Intuitively, this corresponds to picking the y-coordinate of the ROC curve given an x-coordinate.
This metric is a valid choice for realistic scenarios such as large-scale content filtering on social media, where only a small amount of false positives is tolerable.
We consider a fixed false alarm rate of \SI{1}{\percent}.

\paragraph{Evaluating Pre-Trained Detectors}
At first, we test the performance of the pre-trained detectors based on \num{20000} samples, equally divided into real and generated images.
While Wang2020 and Gragnaniello2021 are trained on images from a single \ac{GAN} (ProGAN or StyleGAN2), Mandelli2022 is trained on images from a diverse set of generative models.
The results in the upper half of Table~\ref{tab:pretrained} show that all \ac{GAN}-generated images can be effectively distinguished from real images, with Gragnaniello2021 yielding the best results.
For \ac{DM}-generated, however, the performance of all detectors significantly drops, on average by \SI{15.2}{\percent} \ac{AUROC} compared to \acp{GAN}.
Although the average \ac{AUROC} of \SI{91.4}{\percent} achieved by the best-performing model Gragnaniello2021 (ProGAN variant) appears promising, we argue that in a realistic setting with \SI{1}{\percent} tolerable false positives, detecting only \SI{25.7}{\percent} of all fake images is unacceptable.

To verify that our findings are not limited to our dataset, we extend our evaluation to images from \acp{DM} trained on other datasets, 
variations of ADM, and popular text-to-image models.
We provide details on these additional datasets in \cref{app:dataset_additional}.
The results, given in the lower half of Table~\ref{tab:pretrained}, support the finding that detectors perform significantly worse on \ac{DM}-generated images.
Images from PNDM and LDM trained on LSUN Church are detected better, which we attribute to a dataset-specific bias.
In an extended experiment (see \cref{app:perturbations}) we also find that image perturbations, like compression or blurring, have a stronger adverse effect on the detectability of DM-generated images. 

\begin{figure*}[ht!]
    \centering
    \hspace*{\fill}%
    \subfloat[\ac{AUROC}\label{fig:scratch_auroc}]{%
    \centering
        \includegraphics[scale=\imagescale]{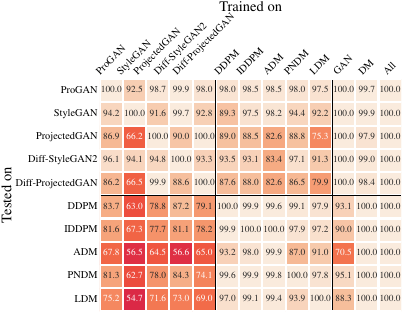}%
    }
    \hspace*{\fill}%
    \subfloat[Pd@1\%FAR]{%
        \centering
        \includegraphics[scale=\imagescale]{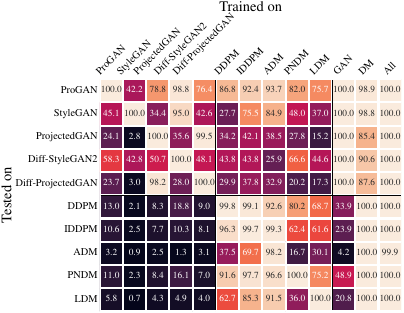}%
    }
    \hspace*{\fill}%
    \caption{{Detection performance for re-trained detectors.} The columns \textit{GAN}, \textit{DM}, and \textit{All} correspond to models trained on samples from all GANs, all DMs, and both, respectively.}
    \label{fig:scratch}
\end{figure*}

\begin{figure*}[ht!]
\centering
    \centering 
    \includegraphics[scale=\imagescale]{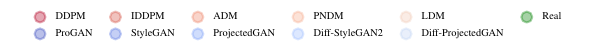}
    
        \hspace*{\fill}%
        \subfloat[Pre-trained \scriptsize{(Blur+JPEG(0.5))}\label{fig:tsne-pre-trained}]{%
            \centering
            \includegraphics[scale=\imagescale]{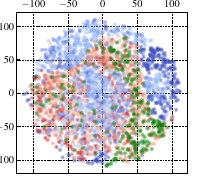}
        }
        \hspace*{\fill}%
        \subfloat[Trained on {\acp{GAN} and \acp{DM}}\label{fig:tsne-All}]{%
            \centering 
            \includegraphics[scale=\imagescale]{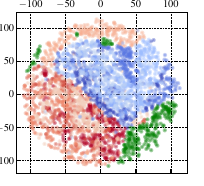}
        }
        \hspace*{\fill}%
        \subfloat[Trained on all \acp{GAN}\label{fig:tsne-GAN}]{%
            \centering 
            \includegraphics[scale=\imagescale]{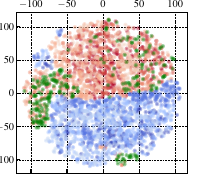}
        }
        \hspace*{\fill}%
        \subfloat[Trained on all \acp{DM}\label{fig:tsne-DM}]{%
            \centering 
            \includegraphics[scale=\imagescale]{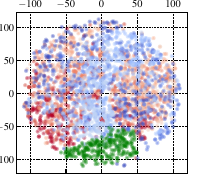}
        }
    \vspace{1mm} %
        \caption{
            {Feature space visualization for the detector Wang2020 via t-SNE of real and generated images in two dimensions.} The features correspond to the representation prior to the last fully-connected layer of the given detector.}
        \label{fig:tsne-small}
\end{figure*}

\paragraph{Generalization of Re-Trained Detectors}
Given the findings presented above, the question arises whether \acp{DM} evade detection in principle, or whether the detection performance can be increased by re-training a detector.\footnote{We carry out the same evaluation with fine-tuned detectors (see \cref{app:finetuning}), leading to very similar results.}
We select the architecture from Wang2020 since the original training code is available and training is relatively efficient.
Furthermore, we choose the configuration Blur+JPEG (0.5) as it yields slightly better scores on average.
For each of the ten generators, we train a detector according to the authors' instructions, using \num{78000} samples for training and \num{2000} samples for validation (equally divided into real (LSUN Bedroom) and fake).
We also consider three aggregated settings in which we train on all images generated by \acp{GAN}, \acp{DM}, and both, respectively.

We report \ac{AUROC} and Pd@1\%FAR for each detector evaluated on all datasets in Figure~\ref{fig:scratch}, based on \num{20000} held-out test samples (\num{10000} real and \num{10000} fake per generator).
All detectors achieve near-perfect scores when evaluated on the dataset they were trained on (represented by the values in the diagonal).
While this is unsurprising for \acp{GAN}, it shows that \acp{DM} \textit{do} exhibit detectable features that a detector can learn.
Regarding generalization, it appears that detectors trained on images from a single \ac{DM} perform better on images from unseen \acp{DM} compared to detectors trained on images from a single \ac{GAN}.
For instance, the detector trained solely on images from ADM achieves a Pd@1\%FAR greater than \SI{90}{\percent} for all other \acp{DM}.
These findings suggest that images generated by \acp{DM} not only contain detectable features, but that these are similar across different architectures and training procedures.

Surprisingly, detectors trained on images from \acp{DM} are significantly more successful in detecting \ac{GAN}-generated images than vice versa.
This becomes most apparent when analyzing the detectors that are trained on all \acp{GAN} and \acp{DM}, respectively.
While the detector trained on images from all \acp{GAN} achieves an average Pd@1\%FAR of \SI{26.34}{\percent} on DM-generated images, the detector trained on images from all \acp{DM} on average detects \SI{94.26}{\percent} of all \ac{GAN}-generated samples.

\begin{figure*}[!ht]
    \centering
    \subfloat[\acp{GAN} \vspace{0.5mm} \label{fig:dft_gan}]{%
        \centering
        \includegraphics[scale=\imagescale]{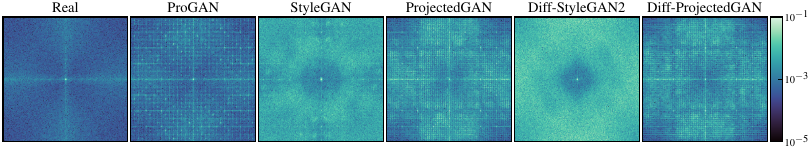}
    }
    \vfill
    \subfloat[\acp{DM}\label{fig:dft_dm}]{%
        \centering
        \includegraphics[scale=\imagescale]{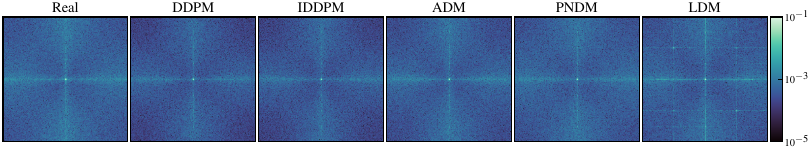}
    }
    \caption{{Mean \acs{DFT} spectrum of real and generated images.} To increase visibility, the color bar is limited to $[10^{-5}, 10^{-1}]$, with values lying outside this interval being clipped.}
    \label{fig:dft}
\end{figure*} 

\paragraph{Analysis of the Learned Feature Spaces}
We conduct a more in-depth analysis of the learned feature spaces to better understand this behavior.
We utilize \mbox{t-SNE}~\cite{maatenVisualizingDataUsing2008} to visualize the extracted features prior to the last fully-connected layer in Figure~\ref{fig:tsne-small}.\footnote{We provide an extended feature space analysis including MMD~\cite{grettonKernelTwoSampleTest2012} in \cref{app:featurespace}. In \cref{app:fakeness} we additionally show example images considered more or less ``fake'' based on the detectors' output scores.}
For the pre-trained Wang2020 we observe a relatively clear separation between real and \ac{GAN}-generated images, while there exists a greater overlap between real and \ac{DM}-generated images (Figure~\ref{fig:tsne-pre-trained}).
These results match the classification results from Table~\ref{tab:pretrained}.
Looking at the detector which is trained on \ac{DM}-generated images only (Figure~\ref{fig:tsne-DM}), the feature representations for \ac{GAN}- and \ac{DM}-generated images appear to be similar.
In contrast, the detectors trained using GAN-generated images or both (Figures~\ref{fig:tsne-GAN} and \ref{fig:tsne-All}) seem to learn distinct feature representations for GAN- and DM-generated images.

Based on these results, we argue that the hypothesis, according to which a detector trained on one family of generative models cannot generalize to a different family~\cite{ojhaUniversalFakeImage2023}, only holds true ``in one direction''.
Given the feature space visualizations, detectors trained on \ac{GAN}-generated images appear to focus mostly on \ac{GAN}-specific artifacts, which may be more prominent and easier to learn.
In contrast, a detector trained exclusively on \ac{DM}-generated images learns a feature representation in which images generated by \acp{GAN} and \acp{DM} are mapped to similar embeddings.
As a consequence, this detector \textit{can} generalize to \ac{GAN}-generated images, since it is not ``distracted'' by family-specific patterns, but learns to detect artifacts which are present in both \ac{GAN}- and \ac{DM}-generated images.

This also implies that \ac{DM}-generated images contain fewer family-specific artifacts.
This becomes apparent when analyzing them in the frequency domain, which we demonstrate in the following section.

\section{\uppercase{Frequency Analysis}}
\label{sec:frequency}
For detecting \ac{GAN}-generated images, exploiting artifacts in the frequency domain has proven to be highly effective~\cite{frankLeveragingFrequencyAnalysis2020}.
Since \acp{DM} contain related building blocks as \acp{GAN} (especially upsampling operations in the underlying U-Net~\cite{ronnebergerUNetConvolutionalNetworks2015}), it seems reasonable to suspect that \ac{DM}-generated exhibit similar artifacts.
In this section, we analyze the spectral properties of \ac{DM}-generated images and compare them to those of \ac{GAN}-generated images.
We investigate potential reasons for the identified frequency characteristics by analyzing the denoising process. 

\paragraph{Transforms}
We use three frequency transforms that have been applied successfully in both traditional image forensics~\cite{lyuNaturalImageStatistics2013} and deepfake detection: \acf{DFT}, \acf{DCT},  and the reduced spectrum~\cite{durallWatchYourUpconvolution2020,dzanicFourierSpectrumDiscrepancies2020,schwarzFrequencyBiasGenerative2021}, which is as a 1D representation of the \ac{DFT}.
While \ac{DFT} and \ac{DCT} visualize frequency artifacts, the reduced spectrum can be used to identify spectrum discrepancies.
The formal definitions of all transforms are provided in \cref{app:transforms}.
The \ac{DCT}-spectra are deferred to \cref{app:dct}, as well as the frequency analyses of the additional datasets (see \cref{app:additional_frequency}).

\paragraph{Analysis of Frequency Artifacts}
\label{sec:properties}
Figure~\ref{fig:dft} depicts the absolute \ac{DFT} spectrum averaged over \num{10000} images from each \ac{GAN} and \ac{DM} trained on LSUN Bedroom.
Before applying the \ac{DFT}, images are transformed to grayscale and, following previous works~\cite{marraGANsLeaveArtificial2019,wangCNNgeneratedImagesAre2020}, high-pass filtered by subtracting a median-filtered version of the image.
For all \acp{GAN} we observe significant artifacts, predominantly in the form of a regular grid, corresponding to previous findings~\cite{zhangDetectingSimulatingArtifacts2019,frankLeveragingFrequencyAnalysis2020}.
In contrast, the \ac{DFT} spectra of images generated by \acp{DM} (see Figure~\ref{fig:dft_dm}), are significantly more similar to the real spectrum with almost no visible artifacts.
LDM is an exception: while being less pronounced than for \acp{GAN}, generated images exhibit a clearly visible grid across their spectrum.
As mentioned in Section~\ref{sec:dataset}, the architecture of LDM differs from the remaining \acp{DM} as the final image is generated using an adversarially trained autoencoder, which could explain the discrepancies.
This observation supports previous findings which suggest that the discriminator is responsible for spectrum deviations~\cite{chenSSDGANMeasuringRealness2021,schwarzFrequencyBiasGenerative2021}.

We conclude that ``traditional'' \acp{DM}, which generate images by gradual denoising, do \textit{not} produce the frequency artifacts known from \acp{GAN}.
Regarding our results in Section~\ref{sec:detection}, this could explain why detectors trained on \ac{GAN} images do not generalize to \acp{DM}, while %
training on \ac{DM}-generated images leads to better generalization.

In a complementary experiment (see \cref{app:logistic_regression}) inspired by~\cite{frankLeveragingFrequencyAnalysis2020} we try to answer whether the reduced amount of frequency artifacts makes DM-generated images less detectable in the frequency domain. On the one hand, using a simple logistic regression classifier, DM-generated images can be better classified in the frequency space than in pixel space. On the other hand, the detection accuracy in both frequency and pixel space is significantly lower compared to GAN-generated images, which makes it difficult to draw conclusions. Nevertheless, the fact that a simple classifier performs significantly better on GAN-generated images strengthens the hypothesis that they exhibit stronger artifacts than images generated by \acp{DM}.

\begin{figure*}
    \centering
    \hspace*{\fill}%
    \subfloat[\acp{GAN}\label{fig:reduced_gan}]{%
        \centering
        \includegraphics[scale=\imagescale]{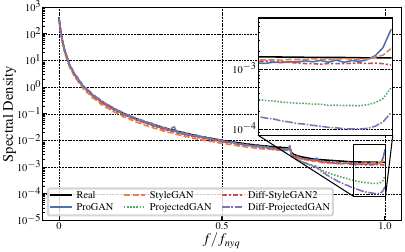}
    }
    \hspace*{\fill}%
    \subfloat[\acp{DM}\label{fig:reduced_dm}]{%
        \centering
        \includegraphics[scale=\imagescale]{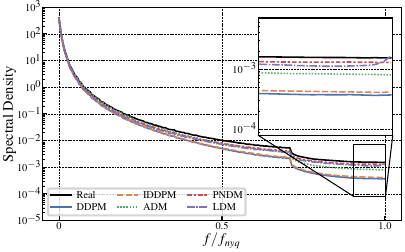}
    }
    \hspace*{\fill}%
    \caption{{Mean reduced spectrum of real and generated images.} The part of the spectrum where \ac{GAN}-characteristic discrepancies occur is magnified.}
    \label{fig:reduced}
\end{figure*}

\paragraph{Analysis of Spectrum Discrepancies}
In a second experiment we analyze how well \acp{GAN} and \acp{DM} are able to reproduce the spectral distribution of real images.
We visualize the reduced spectra for all generators in Figure~\ref{fig:reduced}, again averaged over \num{10000} images.
Except for Diff-StyleGAN2, all \acp{GAN} contain the previously reported elevated high frequencies.
Among the \acp{DM}, these can only be observed for LDM.
This strengthens the hypothesis that it this the autoencoder which causes \ac{GAN}-like frequency characteristics.
However, we observe that all \acp{DM} have a tendency to underestimate the spectral density towards the higher end of the frequency spectrum.
This is particularly noticeable for DDPM, IDDPM, and ADM.

\begin{figure}[hb]
    \centering
    \subfloat[$0 \le t \le 1000$ \vspace{0.5mm}]{%
        \centering
        \includegraphics[scale=\imagescale]{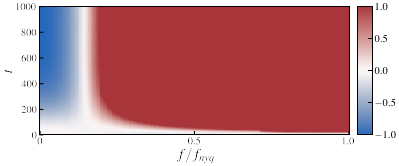}
    }
    \hfill
    \subfloat[$0 \le t \le 100$]{%
        \centering
        \includegraphics[scale=\imagescale]{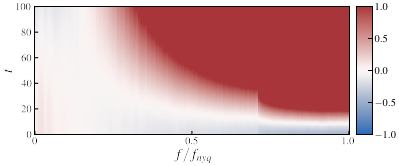}
    }
    \caption{{Spectral density error $\tilde{S}_\text{err}$ throughout the denoising process.} The error is computed relative to the spectrum of real images. We display the error for (a) all sampling steps and (b) a close-up of the last 100 steps. The colorbar is clipped at -1 and 1.}
    \label{fig:denoising}
\end{figure}

\paragraph{Source of Spectrum Underestimation}
Based on these findings, we conduct an additional experiment to identify the source of this spectrum underestimation.
Since \acp{DM} generate images via gradual denoising, we analyze how the spectrum evolves during this denoising process.
For this experiment, we use code and model from ADM~\cite{dhariwalDiffusionModelsBeat2021} trained on LSUN Bedroom.
We generate samples at different time steps $t$ and compare the reduced spectrum (averaged over 512 images) to that of \num{50000} real images.
The results are shown in Figure~\ref{fig:denoising}.

We adopt the figure type from \cite{schwarzFrequencyBiasGenerative2021} and depict the relative spectral density error $\tilde{S}_\text{err} = \tilde{S}_\text{fake} / \tilde{S}_\text{real} - 1$, with the colorbar clipped at \mbox{-1} and 1.
At $t=T=1000$, the image is pure Gaussian noise, which naturally causes strong spectrum deviations.
Around $t=300$, the error starts to decrease, but interestingly it appears that the optimum is not reached at $t=0$, but at $t\approx10$.
It should be noted that while at this step the frequency spectrum is closest to that of real images, they still contain visible noise (see \cref{fig:denoising_examples} in the appendix).
During the final denoising steps, $\tilde{S}_\text{err}$ becomes \textit{negative}, predominantly for higher frequencies, which corresponds to our observations in Figure~\ref{fig:reduced_dm}.
In \cref{app:denoising} we provide additional results, including the spectral density error between the denoising process and the diffusion process at the corresponding step, which further visualizes this behavior.

\begin{figure}[b]
    \centering
    \includegraphics[scale=\imagescale]{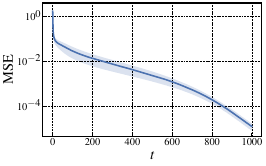}
    \caption{{Mean and standard deviation of the \ac{MSE} for ADM on LSUN Bedroom after training.} The denoising steps towards $t=0$, accounting for high frequencies, have a higher error.}
    \label{fig:recon_loss} 
\end{figure}

We hypothesize that this underestimation towards higher frequencies stems from the learning objective used to train \acp{DM}.
The full line of thoughts builds on observations from research on denoising autoencoders and is given in \cref{app:freq_source}, here we briefly summarize it.
Recalling Section~\ref{sec:background}, \acp{DM} are trained to minimize the \ac{MSE} between the true and predicted noise at different time steps.
The weighting of the \ac{MSE} therefore controls the relative importance of each step.
While the semantic content of an image is generated early during the denoising process, high-frequency details are synthesized near $t=0$~\cite{kingmaVariationalDiffusionModels2021}.
Theoretically, using the variational lower bound $L_\mathrm{vlb}$ as the training objective would yield the highest log-likelihood.
However, training \acp{DM} with $L_\mathrm{vlb}$ is difficult~\cite{hoDenoisingDiffusionProbabilistic2020,nicholImprovedDenoisingDiffusion2021}, which is why in practice modified objectives are used.
The loss proposed in \cite{hoDenoisingDiffusionProbabilistic2020}, $L_\text{simple} = \E_{t,\rvx_0,\epsilon} [\Vert \epsilon - \epsilon_\theta(\rvx_t, t) \Vert^2]$, for example, considers each denoising step as equally important.
Compared to $L_\mathrm{vlb}$, the steps near $t=0$ are significantly down-weighted, trading off a higher perceptual image quality for higher log-likelihood values.
The \ac{MSE} of ADM over $t$ shown in Figure~\ref{fig:recon_loss} demonstrates that the final denoising steps are the most difficult (which is already plain to see as the signal-to-noise ratio increases for $t\to 0$, i.e., the to-be-predicted noise makes up ever smaller fractions of $\rvx_t$).
The hybrid training objective $L_\text{hybrid} = L_\text{simple} + \lambda L_\text{vlb}$~\cite{nicholImprovedDenoisingDiffusion2021}, used in IDDPM and ADM, incorporates $L_\mathrm{vlb}$ (with $\lambda = 0.001$) and already improves upon DDPM in modeling the high-frequency details of an image, but still does not match it accurately.
 
In summary, we conclude that the denoising steps near $t=0$, which govern the high-frequency content of generated images, are the most difficult to model.
By down-weighting the importance of these steps (relatively to the $L_\mathrm{vlb}$), \acp{DM} achieve remarkable perceptual image quality (or benchmark metrics such as FID), but seem to fall short of accurately matching the high-frequency distribution of real data.

\begin{figure}[t]
    \centering
    \includegraphics[scale=1.12]{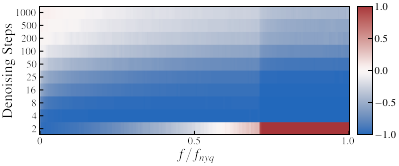}%
    \caption{Spectral density error $\tilde{S}_\text{err}$ for different numbers of denoising steps. The error is computed relatively to the spectrum of real images. The colorbar is clipped at -1 and 1. Note that the $y$-axis is not scaled linearly.}
    \label{fig:evolution_steps}
\end{figure}

\paragraph{Effect of the Number of Sampling Steps}
Lastly, we analyze how the number of sampling steps during the denoising process affects the frequency spectrum.
Previous work reported that increasing the number of steps leads to an improved log-likelihood, corresponding to better reproduction of higher frequencies \cite{nicholImprovedDenoisingDiffusion2021}.
Our results in Figure~\ref{fig:evolution_steps} confirm these findings, increasing the number of denoising steps reduces the underestimation.
In \cref{app:steps} we show that using DDIM~\cite{songDenoisingDiffusionImplicit2022} the spectrum can be reproduced more accurately with fewer steps.
We also analyze the detectability of images generated with different numbers of denoising steps.

\section{\uppercase{Conclusion}}
Deepfakes pose a severe risk for society, and \aclp{DM} have the potential to raise disinformation campaigns to a new level.
Despite the urgency of the problem, research about detecting \ac{DM}-generated images is still in its infancy.
In this work, we provide a much-needed step towards the detection of \ac{DM} deepfakes.
Instead of starting from the ground up, we build on previous achievements in the forensic analysis of \acp{GAN}.
We show that, after re-training, current state-of-the-art detection methods can successfully distinguish real from \ac{DM}-generated images.
Further analysis suggests that \acp{DM} produce fewer detectable artifacts than \acp{GAN}, explaining why detectors trained on \ac{DM}-generated images generalize to \acp{GAN}, but not vice versa.
While artifacts in the frequency domain have been shown to be a characteristic feature of \ac{GAN}-generated images, we find that \acp{DM} predominantly do not have this weakness.
However, we observe a systematic underestimation of the spectral density, which we attribute to the loss function of \acp{DM}.
Whether this mismatch can be exploited for novel detection methods should be part of future research.
We hope that our work can foster the forensic analysis of images generated by \acp{DM} and spark further research towards the effective detection of deepfakes.

\section*{\uppercase{Acknowledgements}}

Funded by the Deutsche Forschungsgemeinschaft (DFG, German Research Foundation) under Germany's Excellence Strategy - EXC 2092 CASA - 390781972.

\bibliographystyle{apalike}
{\small
\bibliography{paper}}

\begin{thebibliography}{}

\bibitem[Chai et~al., 2020]{chaiWhatMakesFake2020}
Chai, L., Bau, D., Lim, S.-N., and Isola, P. (2020).
\newblock What makes fake images detectable? {{Understanding}} properties that generalize.
\newblock In {\em European Conference on Computer Vision (ECCV)}.

\bibitem[Chandrasegaran et~al., 2021]{chandrasegaranCloserLookFourier2021}
Chandrasegaran, K., Tran, N.-T., and Cheung, N.-M. (2021).
\newblock A closer look at {{Fourier}} spectrum discrepancies for {{CNN-generated}} images detection.
\newblock In {\em IEEE Conference on Computer Vision and Pattern Recognition (CVPR)}.

\bibitem[Chen et~al., 2021]{chenSSDGANMeasuringRealness2021}
Chen, Y., Li, G., Jin, C., Liu, S., and Li, T. (2021).
\newblock {{SSD-GAN}}: Measuring the realness in the spatial and spectral domains.
\newblock In {\em AAAI Conference on Artificial Intelligence (AAAI)}.

\bibitem[Choi et~al., 2022]{choiPerceptionPrioritizedTraining2022}
Choi, J., Lee, J., Shin, C., Kim, S., Kim, H., and Yoon, S. (2022).
\newblock Perception prioritized training of diffusion models.
\newblock In {\em IEEE Conference on Computer Vision and Pattern Recognition (CVPR)}.

\bibitem[Corvi et~al., 2023a]{corviIntriguingPropertiesSynthetic2023}
Corvi, R., Cozzolino, D., Poggi, G., Nagano, K., and Verdoliva, L. (2023a).
\newblock Intriguing properties of synthetic images: From generative adversarial networks to diffusion models.
\newblock In {\em IEEE Conference on Computer Vision and Pattern Recognition (CVPR) Workshops}.

\bibitem[Corvi et~al., 2023b]{corviDetectionSyntheticImages2023}
Corvi, R., Cozzolino, D., Zingarini, G., Poggi, G., Nagano, K., and Verdoliva, L. (2023b).
\newblock On the detection of synthetic images generated by diffusion models.
\newblock In {\em IEEE International Conference on Acoustics, Speech and Signal Processing (ICASSP)}.

\bibitem[Cozzolino et~al., 2021]{cozzolinoUniversalGANImage2021}
Cozzolino, D., Gragnaniello, D., Poggi, G., and Verdoliva, L. (2021).
\newblock Towards universal {{GAN}} image detection.
\newblock In {\em International Conference on Visual Communications and Image Processing (VCIP)}.

\bibitem[Dhariwal and Nichol, 2021]{dhariwalDiffusionModelsBeat2021}
Dhariwal, P. and Nichol, A. (2021).
\newblock Diffusion models beat {{GANs}} on image synthesis.
\newblock In {\em Advances in Neural Information Processing Systems (NeurIPS)}.

\bibitem[Dosovitskiy et~al., 2021]{dosovitskiyImageWorth16x162021}
Dosovitskiy, A., Beyer, L., Kolesnikov, A., Weissenborn, D., Zhai, X., Unterthiner, T., Dehghani, M., Minderer, M., Heigold, G., Gelly, S., Uszkoreit, J., and Houlsby, N. (2021).
\newblock An image is worth 16x16 words: Transformers for image recognition at scale.
\newblock {\em International Conference on Learning Representations (ICLR)}.

\bibitem[Durall et~al., 2020]{durallWatchYourUpconvolution2020}
Durall, R., Keuper, M., and Keuper, J. (2020).
\newblock Watch your up-convolution: {{CNN}} based generative deep neural networks are failing to reproduce spectral distributions.
\newblock In {\em IEEE Conference on Computer Vision and Pattern Recognition (CVPR)}.

\bibitem[Dzanic et~al., 2020]{dzanicFourierSpectrumDiscrepancies2020}
Dzanic, T., Shah, K., and Witherden, F. (2020).
\newblock Fourier spectrum discrepancies in deep network generated images.
\newblock In {\em Advances in Neural Information Processing Systems (NeurIPS)}.

\bibitem[Farid, 2022a]{faridLightingConsistencyPaint2022}
Farid, H. (2022a).
\newblock Lighting (in)consistency of paint by text.
\newblock {\em arXiv preprint}.

\bibitem[Farid, 2022b]{faridPerspectiveConsistencyPaint2022}
Farid, H. (2022b).
\newblock Perspective (in)consistency of paint by text.
\newblock {\em arXiv preprint}.

\bibitem[Frank et~al., 2020]{frankLeveragingFrequencyAnalysis2020}
Frank, J., Eisenhofer, T., Sch{\"o}nherr, L., Fischer, A., Kolossa, D., and Holz, T. (2020).
\newblock Leveraging frequency analysis for deep fake image recognition.
\newblock In {\em International Conference on Machine Learning (ICML)}.

\bibitem[Geras and Sutton, 2015]{gerasScheduledDenoisingAutoencoders2015}
Geras, K.~J. and Sutton, C. (2015).
\newblock Scheduled denoising autoencoders.
\newblock In {\em International Conference on Learning Representations (ICLR)}.

\bibitem[Geras and Sutton, 2016]{gerasCompositeDenoisingAutoencoders2016}
Geras, K.~J. and Sutton, C. (2016).
\newblock Composite denoising autoencoders.
\newblock In {\em European Conference on Machine Learning and Knowledge Discovery in Databases (ECML PKDD)}.

\bibitem[Girish et~al., 2021]{girishDiscoveryAttributionOpenworld2021}
Girish, S., Suri, S., Rambhatla, S.~S., and Shrivastava, A. (2021).
\newblock Towards discovery and attribution of open-world {{GAN}} generated images.
\newblock In {\em IEEE Conference on Computer Vision and Pattern Recognition (CVPR)}.

\bibitem[Gragnaniello et~al., 2021]{gragnanielloAreGANGenerated2021}
Gragnaniello, D., Cozzolino, D., Marra, F., Poggi, G., and Verdoliva, L. (2021).
\newblock Are {{GAN}} generated images easy to detect? {{A}} critical analysis of the state-of-the-art.
\newblock In {\em IEEE International Conference on Multimedia and Expo (ICME)}.

\bibitem[Gretton et~al., 2012]{grettonKernelTwoSampleTest2012}
Gretton, A., Borgwardt, K.~M., Rasch, M.~J., Sch{\"o}lkopf, B., and Smola, A. (2012).
\newblock A kernel two-sample test.
\newblock {\em Journal of Machine Learning Research (JMLR)}, 13(25):723--773.

\bibitem[He et~al., 2016]{heDeepResidualLearning2016}
He, K., Zhang, X., Ren, S., and Sun, J. (2016).
\newblock Deep residual learning for image recognition.
\newblock In {\em IEEE Conference on Computer Vision and Pattern Recognition (CVPR)}.

\bibitem[Heusel et~al., 2017]{heuselGANsTrainedTwo2017}
Heusel, M., Ramsauer, H., Unterthiner, T., Nessler, B., and Hochreiter, S. (2017).
\newblock {{GANs}} trained by a two time-scale update rule converge to a local nash equilibrium.
\newblock In {\em Advances in Neural Information Processing Systems (NeurIPS)}.

\bibitem[Ho et~al., 2020]{hoDenoisingDiffusionProbabilistic2020}
Ho, J., Jain, A., and Abbeel, P. (2020).
\newblock Denoising diffusion probabilistic models.
\newblock In {\em Advances in Neural Information Processing Systems (NeurIPS)}.

\bibitem[Hu et~al., 2021]{huExposingGANGeneratedFaces2021}
Hu, S., Li, Y., and Lyu, S. (2021).
\newblock Exposing {{GAN-Generated}} faces using inconsistent corneal specular highlights.
\newblock In {\em IEEE International Conference on Acoustics, Speech and Signal Processing (ICASSP)}.

\bibitem[Huang, 2023]{huangWhyPopeFrancis2023}
Huang, K. (2023).
\newblock Why {{Pope Francis}} is the star of {{A}}.{{I}}.-generated photos.
\newblock {\em The New York Times}.

\bibitem[Hulzebosch et~al., 2020]{hulzeboschDetectingCNNgeneratedFacial2020}
Hulzebosch, N., Ibrahimi, S., and Worring, M. (2020).
\newblock Detecting {{CNN-generated}} facial images in real-world scenarios.
\newblock In {\em IEEE Conference on Computer Vision and Pattern Recognition (CVPR) Workshops}.

\bibitem[Jeong et~al., 2022]{jeongFingerprintNet2022}
Jeong, Y., Kim, D., Ro, Y., Kim, P., and Choi, J. (2022).
\newblock {{FingerprintNet}}: Synthesized fingerprints for generated image detection.
\newblock In {\em European Conference on Computer Vision (ECCV)}.

\bibitem[Karras et~al., 2018]{karrasProgressiveGrowingGANs2018}
Karras, T., Aila, T., Laine, S., and Lehtinen, J. (2018).
\newblock Progressive growing of {{GANs}} for improved quality, stability, and variation.
\newblock In {\em International Conference on Learning Representations (ICLR)}.

\bibitem[Karras et~al., 2019]{karrasStylebasedGeneratorArchitecture2019}
Karras, T., Laine, S., and Aila, T. (2019).
\newblock A style-based generator architecture for generative adversarial networks.
\newblock In {\em IEEE Conference on Computer Vision and Pattern Recognition (CVPR)}.

\bibitem[Karras et~al., 2020]{karrasAnalyzingImprovingImage2020}
Karras, T., Laine, S., Aittala, M., Hellsten, J., Lehtinen, J., and Aila, T. (2020).
\newblock Analyzing and improving the image quality of {{StyleGAN}}.
\newblock In {\em IEEE Conference on Computer Vision and Pattern Recognition (CVPR)}.

\bibitem[Khayatkhoei and Elgammal, 2022]{khayatkhoeiSpatialFrequencyBias2022}
Khayatkhoei, M. and Elgammal, A. (2022).
\newblock Spatial frequency bias in convolutional generative adversarial networks.
\newblock {\em AAAI Conference on Artificial Intelligence (AAAI)}.

\bibitem[Kingma et~al., 2021]{kingmaVariationalDiffusionModels2021}
Kingma, D., Salimans, T., Poole, B., and Ho, J. (2021).
\newblock Variational diffusion models.
\newblock In {\em Advances in Neural Information Processing Systems (NeurIPS)}.

\bibitem[Kingma and Gao, 2023]{kingma2023understanding}
Kingma, D.~P. and Gao, R. (2023).
\newblock Understanding diffusion objectives as the {ELBO} with simple data augmentation.
\newblock In {\em Advances in Neural Information Processing Systems (NeurIPS)}.

\bibitem[Liu et~al., 2022]{liuPseudoNumericalMethods2022}
Liu, L., Ren, Y., Lin, Z., and Zhao, Z. (2022).
\newblock Pseudo numerical methods for diffusion models on manifolds.
\newblock In {\em International Conference on Learning Representations (ICLR)}.

\bibitem[Liu et~al., 2020]{liuGlobalTextureEnhancement2020}
Liu, Z., Qi, X., and Torr, P. H.~S. (2020).
\newblock Global texture enhancement for fake face detection in the wild.
\newblock In {\em IEEE Conference on Computer Vision and Pattern Recognition (CVPR)}.

\bibitem[Lyu, 2008]{lyuNaturalImageStatistics2013}
Lyu, S. (2008).
\newblock {\em Natural Image Statistics in Digital Image Forensics}.
\newblock PhD thesis, Dartmouth College.

\bibitem[Mandelli et~al., 2022]{mandelliDetectingGANgeneratedImages2022}
Mandelli, S., Bonettini, N., Bestagini, P., and Tubaro, S. (2022).
\newblock Detecting {GAN}-generated images by orthogonal training of multiple {CNNs}.
\newblock In {\em IEEE International Conference on Image Processing (ICIP)}.

\bibitem[Marra et~al., 2019]{marraGANsLeaveArtificial2019}
Marra, F., Gragnaniello, D., Verdoliva, L., and Poggi, G. (2019).
\newblock Do {{GANs}} leave artificial fingerprints?
\newblock In {\em IEEE Conference on Multimedia Information Processing and Retrieval (MIPR)}.

\bibitem[McCloskey and Albright, 2019]{mccloskeyDetectingGANgeneratedImagery2019}
McCloskey, S. and Albright, M. (2019).
\newblock Detecting {{GAN-generated}} imagery using saturation cues.
\newblock In {\em IEEE International Conference on Image Processing (ICIP)}.

\bibitem[Nataraj et~al., 2019]{natarajDetectingGANGenerated2019}
Nataraj, L., Mohammed, T.~M., Manjunath, B.~S., Chandrasekaran, S., Flenner, A., Bappy, J.~H., and {Roy-Chowdhury}, A.~K. (2019).
\newblock Detecting {{GAN}} generated fake images using co-occurrence matrices.
\newblock {\em Electronic Imaging}.

\bibitem[Nichol and Dhariwal, 2021]{nicholImprovedDenoisingDiffusion2021}
Nichol, A.~Q. and Dhariwal, P. (2021).
\newblock Improved denoising diffusion probabilistic models.
\newblock In {\em International Conference on Machine Learning (ICML)}.

\bibitem[Nightingale and Farid, 2022]{nightingaleAIsynthesizedFacesAre2022}
Nightingale, S.~J. and Farid, H. (2022).
\newblock {{AI-synthesized}} faces are indistinguishable from real faces and more trustworthy.
\newblock {\em Proceedings of the National Academy of Sciences}.

\bibitem[Ojha et~al., 2023]{ojhaUniversalFakeImage2023}
Ojha, U., Li, Y., and Lee, Y.~J. (2023).
\newblock Towards universal fake image detectors that generalize across generative models.
\newblock In {\em IEEE Conference on Computer Vision and Pattern Recognition (CVPR)}.

\bibitem[Radford et~al., 2021]{radfordLearningTransferableVisual2021}
Radford, A., Kim, J.~W., Hallacy, C., Ramesh, A., Goh, G., Agarwal, S., Sastry, G., Askell, A., Mishkin, P., Clark, J., Krueger, G., and Sutskever, I. (2021).
\newblock Learning transferable visual models from natural language supervision.
\newblock In {\em International Conference on Machine Learning (ICML)}.

\bibitem[Ramesh et~al., 2022]{rameshHierarchicalTextconditionalImage2022}
Ramesh, A., Dhariwal, P., Nichol, A., Chu, C., and Chen, M. (2022).
\newblock Hierarchical text-conditional image generation with {{CLIP}} latents.
\newblock {\em arXiv preprint}.

\bibitem[Rissanen et~al., 2023]{rissanenGenerativeModellingInverse2022}
Rissanen, S., Heinonen, M., and Solin, A. (2023).
\newblock Generative modelling with inverse heat dissipation.
\newblock In {\em International Conference on Learning Representations (ICLR)}.

\bibitem[Rombach et~al., 2022]{rombachHighresolutionImageSynthesis2022}
Rombach, R., Blattmann, A., Lorenz, D., Esser, P., and Ommer, B. (2022).
\newblock High-resolution image synthesis with latent diffusion models.
\newblock In {\em IEEE Conference on Computer Vision and Pattern Recognition (CVPR)}.

\bibitem[Ronneberger et~al., 2015]{ronnebergerUNetConvolutionalNetworks2015}
Ronneberger, O., Fischer, P., and Brox, T. (2015).
\newblock U-{{Net}}: Convolutional networks for biomedical image segmentation.
\newblock In {\em International Conference on Medical Image Computing and Computer Assisted Intervention}.

\bibitem[Russakovsky et~al., 2015]{russakovskyImageNetLargeScale2015}
Russakovsky, O., Deng, J., Su, H., Krause, J., Satheesh, S., Ma, S., Huang, Z., Karpathy, A., Khosla, A., Bernstein, M., Berg, A.~C., and {Fei-Fei}, L. (2015).
\newblock {{ImageNet}} large scale visual recognition challenge.
\newblock {\em International Journal of Computer Vision (IJCV)}.

\bibitem[Saharia et~al., 2022]{sahariaPhotorealisticTexttoimageDiffusion2022}
Saharia, C., Chan, W., Saxena, S., Li, L., Whang, J., Denton, E., Ghasemipour, S. K.~S., Ayan, B.~K., Mahdavi, S.~S., Lopes, R.~G., Salimans, T., Ho, J., Fleet, D.~J., and Norouzi, M. (2022).
\newblock Photorealistic text-to-image diffusion models with deep language understanding.
\newblock {\em arXiv preprint}.

\bibitem[Salimans and Ho, 2022]{salimansProgressiveDistillationFast2022}
Salimans, T. and Ho, J. (2022).
\newblock Progressive distillation for fast sampling of diffusion models.
\newblock In {\em International Conference on Learning Representations (ICLR)}.

\bibitem[Sauer et~al., 2021]{sauerProjectedGANsConverge2021}
Sauer, A., Chitta, K., M{\"u}ller, J., and Geiger, A. (2021).
\newblock Projected {{GANs}} converge faster.
\newblock In {\em Advances in Neural Information Processing Systems (NeurIPS)}.

\bibitem[Schwarz et~al., 2021]{schwarzFrequencyBiasGenerative2021}
Schwarz, K., Liao, Y., and Geiger, A. (2021).
\newblock On the frequency bias of generative models.
\newblock In {\em Advances in Neural Information Processing Systems (NeurIPS)}.

\bibitem[Sha et~al., 2023]{shaDEFAKEDetectionAttribution2022}
Sha, Z., Li, Z., Yu, N., and Zhang, Y. (2023).
\newblock {{DE-FAKE}}: Detection and attribution of fake images generated by text-to-image diffusion models.
\newblock {\em ACM SIGSAC Conference on Computer and Communications Security (CCS)}.

\bibitem[{Sohl-Dickstein} et~al., 2015]{sohl-dicksteinDeepUnsupervisedLearning2015}
{Sohl-Dickstein}, J., Weiss, E., Maheswaranathan, N., and Ganguli, S. (2015).
\newblock Deep unsupervised learning using nonequilibrium thermodynamics.
\newblock In {\em International Conference on Machine Learning (ICML)}.

\bibitem[Song et~al., 2022a]{songDenoisingDiffusionImplicit2022}
Song, J., Meng, C., and Ermon, S. (2022a).
\newblock Denoising diffusion implicit models.
\newblock In {\em International Conference on Learning Representations (ICLR)}.

\bibitem[Song and Ermon, 2019]{songGenerativeModelingEstimating2019}
Song, Y. and Ermon, S. (2019).
\newblock Generative modeling by estimating gradients of the data distribution.
\newblock In {\em Advances in Neural Information Processing Systems (NeurIPS)}.

\bibitem[Song and Ermon, 2020]{songImprovedTechniquesTraining2020}
Song, Y. and Ermon, S. (2020).
\newblock Improved techniques for training score-based generative models.
\newblock In {\em Advances in Neural Information Processing Systems (NeurIPS)}.

\bibitem[Song et~al., 2022b]{songScorebasedGenerativeModeling2022}
Song, Y., {Sohl-Dickstein}, J., Kingma, D.~P., Kumar, A., Ermon, S., and Poole, B. (2022b).
\newblock Score-based generative modeling through stochastic differential equations.
\newblock In {\em International Conference on Learning Representations (ICLR)}.

\bibitem[Tan and Le, 2019]{tanEfficientNetRethinkingModel2019}
Tan, M. and Le, Q. (2019).
\newblock {{EfficientNet}}: Rethinking model scaling for convolutional neural networks.
\newblock In {\em International Conference on Machine Learning (ICML)}.

\bibitem[van~der Maaten and Hinton, 2008]{maatenVisualizingDataUsing2008}
van~der Maaten, L. and Hinton, G. (2008).
\newblock Visualizing data using t-{{SNE}}.
\newblock {\em Journal of Machine Learning Research (JMLR)}.

\bibitem[Verdoliva, 2020]{verdolivaMediaForensicsDeepFakes2020}
Verdoliva, L. (2020).
\newblock Media forensics and {{DeepFakes}}: {{An}} overview.
\newblock {\em IEEE Journal of Selected Topics in Signal Processing}.

\bibitem[Vincent et~al., 2010]{vincentStackedDenoisingAutoencoders2010}
Vincent, P., Larochelle, H., Lajoie, I., Bengio, Y., and Manzagol, P.-A. (2010).
\newblock Stacked denoising autoencoders: Learning useful representations in a deep network with a local denoising criterion.
\newblock {\em Journal of Machine Learning Research (JMLR)}.

\bibitem[Wallace, 1991]{wallaceJPEGStillPicture1991}
Wallace, G.~K. (1991).
\newblock The {{JPEG}} still picture compression standard.
\newblock {\em Communications of the ACM}.

\bibitem[Wang et~al., 2020]{wangCNNgeneratedImagesAre2020}
Wang, S.-Y., Wang, O., Zhang, R., Owens, A., and Efros, A.~A. (2020).
\newblock {{CNN-generated}} images are surprisingly easy to spot... for now.
\newblock In {\em IEEE Conference on Computer Vision and Pattern Recognition (CVPR)}.

\bibitem[Wang et~al., 2023]{wangDIREDiffusiongeneratedImage2023}
Wang, Z., Bao, J., Zhou, W., Wang, W., Hu, H., Chen, H., and Li, H. (2023).
\newblock {{DIRE}} for diffusion-generated image detection.
\newblock {\em IEEE International Conference on Computer Vision (ICCV)}.

\bibitem[Wang et~al., 2022a]{wangDiffusionGANTrainingGANs2022}
Wang, Z., Zheng, H., He, P., Chen, W., and Zhou, M. (2022a).
\newblock Diffusion-{{GAN}}: Training {GANs} with diffusion.
\newblock {\em arXiv preprint}.

\bibitem[Wang et~al., 2022b]{wang2022diffusiondb}
Wang, Z.~J., Montoya, E., Munechika, D., Yang, H., Hoover, B., and Chau, D.~H. (2022b).
\newblock {{DiffusionDB}}: A large-scale prompt gallery dataset for text-to-image generative models.
\newblock {\em arXiv preprint}.

\bibitem[Xiao et~al., 2022]{xiaoTacklingGenerativeLearning2022}
Xiao, Z., Kreis, K., and Vahdat, A. (2022).
\newblock Tackling the generative learning trilemma with denoising diffusion {{GANs}}.
\newblock In {\em International Conference on Learning Representations (ICLR)}.

\bibitem[Xuan et~al., 2019]{xuanGeneralizationGANImage2019}
Xuan, X., Peng, B., Wang, W., and Dong, J. (2019).
\newblock On the generalization of {{GAN}} image forensics.
\newblock In {\em Biometric Recognition (CCBR)}.

\bibitem[Yang et~al., 2023]{yang2023DiffusionReview}
Yang, L., Zhang, Z., Song, Y., Hong, S., Xu, R., Zhao, Y., Zhang, W., Cui, B., and Yang, M.-H. (2023).
\newblock Diffusion models: A comprehensive survey of methods and applications.
\newblock {\em ACM Computing Surveys}.

\bibitem[Yu et~al., 2016]{yuLSUNConstructionLargescale2016}
Yu, F., Seff, A., Zhang, Y., Song, S., Funkhouser, T., and Xiao, J. (2016).
\newblock {{LSUN}}: Construction of a large-scale image dataset using deep learning with humans in the loop.
\newblock {\em arXiv preprint}.

\bibitem[Zhang et~al., 2019]{zhangDetectingSimulatingArtifacts2019}
Zhang, X., Karaman, S., and Chang, S.-F. (2019).
\newblock Detecting and simulating artifacts in {{GAN}} fake images.
\newblock In {\em IEEE International Workshop on Information Forensics and Security (WIFS)}.

\end{thebibliography}

\onecolumn

\section*{\uppercase{Appendix}}

\appendix

\section{\uppercase{Data Collection}}
\subsection{LSUN Bedroom}
\label{app:dataset}
\begin{figure*}[ht!]
    \centering
    \includegraphics[scale=\imagescale]{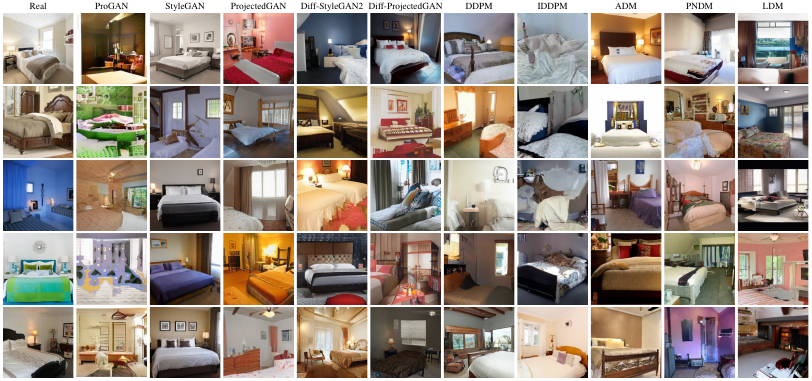}
    \caption{{Non-curated example images for real LSUN Bedroom, \ac{GAN}-generated, and \ac{DM}-generated images.}}
    \label{fig:examples}
\end{figure*}

\paragraph{LSUN Bedroom \protect\cite{yuLSUNConstructionLargescale2016}}
We download and extract the \texttt{lmbd} database files using the official repository\footnote{\url{https://github.com/fyu/lsun}}.
The images are center-cropped to 256$\times$256 pixels.

\paragraph{ProGAN \protect\cite{karrasProgressiveGrowingGANs2018}}
\hspace{-2.5mm}
We download the first \num{10000} samples from the non-curated collection provided by the authors.\footnote{\url{https://github.com/tkarras/progressive_growing_of_gans}}

\paragraph{StyleGAN \protect\cite{karrasStylebasedGeneratorArchitecture2019}}
We download the first \num{10000} samples generated with $\psi=0.5$ from the non-curated collection provided by the authors.\footnote{\url{https://github.com/NVlabs/stylegan}}

\paragraph{ProjectedGAN \protect\cite{sauerProjectedGANsConverge2021}}
We sample \num{10000} images using code and pre-trained models provided by the authors using the default configuration (\texttt{--trunc=1.0}).\footnote{\url{https://github.com/autonomousvision/projected_gan}}

\paragraph{Diff-StyleGAN2 and Diff-ProjectedGAN \protect\cite{wangDiffusionGANTrainingGANs2022}}
We sample \num{10000} images using code and pre-trained models provided by the authors using the default configuration.\footnote{\url{https://github.com/Zhendong-Wang/Diffusion-GAN}}

\paragraph{DDPM \protect\cite{hoDenoisingDiffusionProbabilistic2020}, IDDPM \protect\cite{nicholImprovedDenoisingDiffusion2021}, and ADM \protect\cite{dhariwalDiffusionModelsBeat2021}}
We download the samples provided by the authors of ADM\footnote{\url{https://github.com/openai/guided-diffusion}} and extract the first \num{10000} samples for each generator.
For ADM on LSUN, we select the models trained with dropout.

\paragraph{PNDM \protect\cite{liuPseudoNumericalMethods2022}}
We sample \num{10000} images using code and pre-trained model provided by the authors.\footnote{\url{https://github.com/luping-liu/PNDM}}
We specify \texttt{--method F-PNDM} and \texttt{--sample\_speed 20} for LSUN Bedroom and \texttt{--sample\_speed 10} for LSUN Church, as these are the settings leading to the lowest \ac{FID} according to Tables 5 and 6 in the original publication.

\paragraph{LDM \protect\cite{rombachHighresolutionImageSynthesis2022}}
We sample \num{10000} images using code and pre-trained models provided by the authors using settings from the corresponding table in the repository.\footnote{\url{https://github.com/CompVis/latent-diffusion}}
For LSUN Church there is an inconsistency between the repository and the paper, we choose 200 DDIM steps (\texttt{-c 200}) as reported in the paper.

\subsection{Additional Datasets}
\label{app:dataset_additional}
Here we provide details on the additional datasets analyzed in Table~\ref{tab:pretrained}.
Note that ADM-G-U refers to the two-stage up-sampling stack in which images are generated at a resolution of 64$\times$64 and subsequently up-sampled to 256$\times$256 pixels using a second model \cite{dhariwalDiffusionModelsBeat2021}.
The generated images are obtained according to the instructions given in the previous section.

Due to the relevance of facial images in the context of deepfakes, we also include two \acp{DM} not yet considered, P2 and ADM' \cite{choiPerceptionPrioritizedTraining2022}, trained on FFHQ \cite{karrasStylebasedGeneratorArchitecture2019}.
ADM' is a smaller version of ADM with 93 million instead of more than 500 million parameters.\footnote{\url{https://github.com/jychoi118/P2-weighting\#training-your-models}}
P2 is similar to ADM' but features a modified weighting scheme which improves performance by assigning higher weights to diffusion steps where perceptually rich contents are learned \cite{choiPerceptionPrioritizedTraining2022}.
We download checkpoints for both models from the official repository and sample images according to the authors' instructions.

Real images from LSUN \cite{yuLSUNConstructionLargescale2016}, ImageNet \cite{russakovskyImageNetLargeScale2015}, and FFHQ \cite{karrasStylebasedGeneratorArchitecture2019} are downloaded from their official sources.
Images from LSUN Cat/Horse, FFHQ, and ImageNet are resized and cropped to 256$\times$256 pixels by applying the same pre-processing that was used when preparing the training data for the model they are compared against.
For all datasets we collect \num{10000} real and \num{10000} generated images.

Images from Stable Diffusion\footnote{\url{https://stability.ai/blog/stable-diffusion-public-release}} are generated using the \textit{diffusers} library\footnote{\url{https://huggingface.co/docs/diffusers/index}} with default settings.
For each version, we generate \num{10000} images using prompts from DiffusionDB~\cite{wang2022diffusiondb}.
Since Midjourney\footnote{\url{https://www.midjourney.com}} is proprietary, we collect 300 images created using the ``--v 5'' flag from the official Discord server.
As real images, we take a subset of \num{10000} images from LAION-Aesthetics V2\footnote{\url{https://laion.ai/blog/laion-aesthetics/}} with aesthetics scores greater than 6.5.
For the detection experiments, we use the entire images, for computing frequency spectra we take center crops of size 256$\times$256.

\clearpage

\section{\uppercase{Supplemental Material on the Detection Analysis}}
\subsection{Descriptions of Detectors}
\label{app:detectors}
\paragraph{Wang2020~\protect\cite{wangCNNgeneratedImagesAre2020}}
In this influential work, the authors demonstrate that a standard deep \ac{CNN} trained on data from a single generator performs surprisingly well on unseen images.
They train a ResNet-50 \cite{heDeepResidualLearning2016} on \num{720000} images from 20 LSUN \cite{yuLSUNConstructionLargescale2016} categories (not including Bedroom and Church), equally divided into real images and images generated by ProGAN \cite{karrasProgressiveGrowingGANs2018}.
The trained binary classifier is able to distinguish real from generated images from a variety of generative models and datasets.
The authors further show that extensive data augmentation in the form of random flipping, blurring, and JPEG compression generally improves generalization.
Moreover, the authors provide two pre-trained model configurations, Blur+JPEG (0.1) and Blur+JPEG (0.5), where the value in parentheses denotes the probability of blurring and JPEG compression, respectively.
They achieve an average precision of \SI{92.6}{\percent} and \SI{90.8}{\percent}, respectively.
The work suggests that CNN-generated images contain common artifacts (or fingerprints) which make them distinguishable from real images.

\paragraph{Gragnaniello2021~\protect\cite{gragnanielloAreGANGenerated2021}}
Building upon the architecture and dataset from \cite{wangCNNgeneratedImagesAre2020}, the authors of this work experiment with different variations to further improve the detection performance in real-world scenarios.
The most promising variant, \textit{no-down}, removes downsampling from the first layer, increasing the average accuracy from \SI{80.71}{\percent} to \SI{94.42}{\percent}, at the cost of more trainable parameters.
They also train a model with the same architecture on images generated by StyleGAN2 \cite{karrasAnalyzingImprovingImage2020} instead of ProGAN, which further improves the accuracy to \SI{98.48}{\percent}.

\paragraph{Mandelli2022~\protect\cite{mandelliDetectingGANgeneratedImages2022}}
Unlike the other two methods, this work uses an ensemble of five orthogonal \acp{CNN} to detect fake images not seen during training.
All \acp{CNN} are based on the EfficientNet-B4 model \cite{tanEfficientNetRethinkingModel2019} but are trained on different datasets.
Dataset orthogonality refers to images having different content, different processing (e.g., JPEG compression), or being generated by different \acp{GAN}.
They argue that by having different datasets, each \ac{CNN} learns to detect different characteristics of real and generated images, improving the overall performance and generalization ability.
For each \ac{CNN}, a score is computed for $\approx$ 200 random patches extracted from the image.
These scores are combined using a novel patch aggregation strategy which assumes that an image is fake if at least one patch is classified as being fake.
Finally, the output score is computed by averaging the individual scores of all five \acp{CNN}.
It should be noted that, besides several \acp{GAN}, the training dataset also includes samples generated by a score-based model~\cite{songScorebasedGenerativeModeling2022}.

\subsection{Detection Results for Fine-Tuned Classifiers}
\label{app:finetuning}
\begin{figure*}[!ht]
\vspace{-2mm}
    \centering
    \hspace*{\fill}%
    \subfloat[\ac{AUROC}\label{fig:finetuned_auroc}]{%
        \includegraphics[scale=\imagescale]{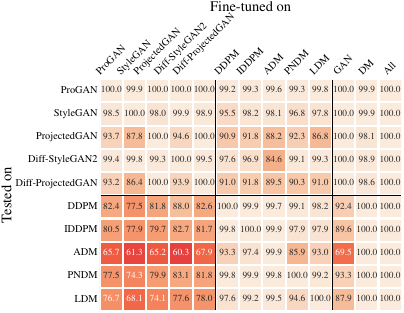}%
    }
    \hspace*{\fill}%
    \subfloat[Pd@1\%\label{fig:finetuned_pd1}]{%
        \includegraphics[scale=\imagescale]{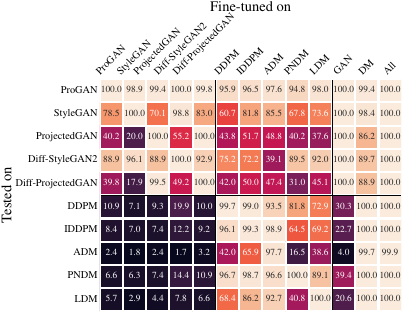}%
    }
    \hspace*{\fill}%
    \caption{{Detection performance for fine-tuned detectors.} The columns \textit{GAN}, \textit{DM}, and \textit{All} correspond to models trained on samples from all GANs, all DMs, and both, respectively.}
    \label{fig:finetuned}
    \vspace{-9mm}
\end{figure*}

We repeat the experiment used to generated Figure~\ref{fig:scratch} but fine-tune each classifier instead of re-training it.
The results are shown in Figure~\ref{fig:finetuned}.
For most constellations of training and test data, the results are similar compared to those in Figure~\ref{fig:scratch}.
We suspect that during fine-tuning the existing knowledge is mostly ``overwritten''.

\clearpage

\subsection{Extended Feature Space Analysis}
\label{app:featurespace}
We extend the feature space analysis conducted in Section~\ref{sec:detection} by estimating the distance between the distributions of real and generated images in feature space using maximum mean discrepancy (MMD, \cite{grettonKernelTwoSampleTest2012}).
The $2048$-dimensional features are extracted prior to the last fully-connected layer of the detectors Wang2020 and Gragnaniello2021.
We employ a Gaussian kernel with $\sigma$ as the median distance between instances in the combined sample as suggested by \cite{grettonKernelTwoSampleTest2012} and calculate the MMD between the representations of \num{10000} generated and \num{10000} real images.

\begin{figure}[ht]
    \centering
    \includegraphics[scale=\imagescale]{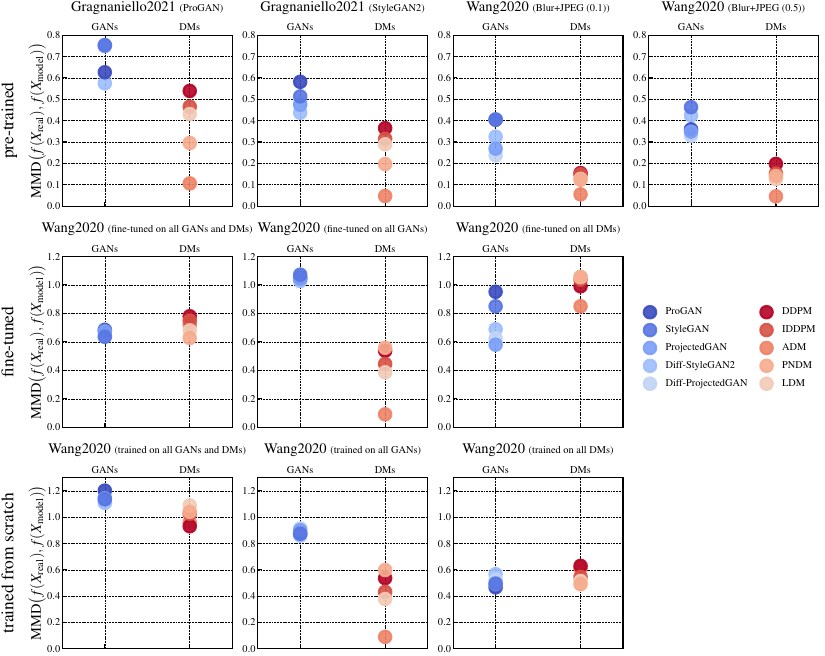}
    \caption{Feature space visualization for different detectors (pre-trained, fine-tuned and trained from scratch) via MMD of real (LSUN Bedroom) and generated images. The features $f(\cdot)$ correspond to the representation prior to the last fully-connected layer of the respective detection method.}
    \label{fig:MMD}
\end{figure}

We consider three sets of detection methods, namely pre-trained, fine-tuned and re-trained from scratch.
Starting with pre-trained detection methods (top row in Figure~\ref{fig:MMD}), we observe that the MMD between representations of DM-generated images and those of real images are considerably lower in comparison to GAN-generated images. 
With the results from Table~\ref{tab:pretrained} in mind, we can conclude that the pre-trained detectors extract features that allow to reliably separate GAN-generated images from real images, while these features are less informative to spot DM-generated images.

When fine-tuning/re-training on generated images from both, DMs and GANs, the MMDs in feature space are roughly on par for both model sets (middle and bottom row in Figure~\ref{fig:MMD}).
When fine-tuning/training solely on images from one model class we observe an imbalance:
Detectors fine-tuned or trained on images from DMs are able to achieve relatively higher MMDs for GAN-generated images than vice versa, supporting our previous findings.

\clearpage

\begin{figure}[H]
    \centering
    \includegraphics[scale=\imagescale]{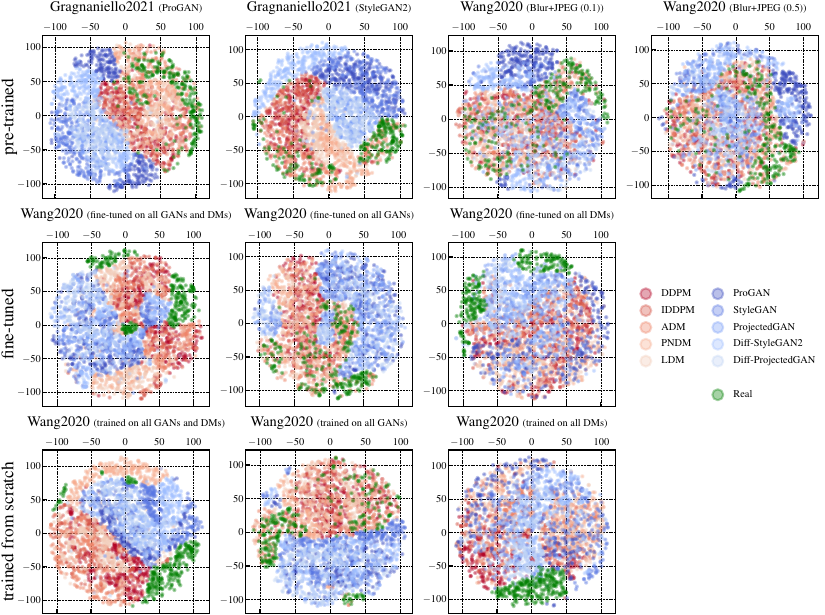}
    \caption{Feature space visualization for different detectors (pre-trained, fine-tuned and trained from scratch) via t-SNE of real (LSUN Bedroom) and generated images in two dimensions. The features correspond to the representation prior to the last fully-connected layer of the respective detection method.}
    \label{fig:tsne}
    \vspace{+2mm}
\end{figure}

For completeness, we provide the t-SNE visualizations~\cite{maatenVisualizingDataUsing2008} analogous to Figure~\ref{fig:MMD} in Figure~\ref{fig:tsne}.
We use the scikit-learn implementation of t-SNE with the default settings for all \num{110000} images in our dataset (\num{10000} images per GAN/DM and \num{10000} real images) and visualize 250 images per class.

\subsection{Effect of Image Perturbations}
\label{app:perturbations}
In most real-world scenarios, like uploading to social media, images are processed, which is why several previous works consider common image perturbations when evaluating detectors \cite{wangCNNgeneratedImagesAre2020,hulzeboschDetectingCNNgeneratedFacial2020,liuGlobalTextureEnhancement2020,frankLeveragingFrequencyAnalysis2020}.
We follow the protocol of \cite{frankLeveragingFrequencyAnalysis2020} and apply blurring using a Gaussian filter (kernel size sampled from $\{3,5,7,9\}$, cropping with subsequent upsampling (crop factor sampled from $U(5, 20)$), JPEG compression (quality factor sampled from $U(10, 75)$), and Gaussian noising (variance sampled from $U(5, 20)$).
Unlike \cite{frankLeveragingFrequencyAnalysis2020}, we apply each perturbation with a probability of \SI{100}{\percent} to study its effect on the detection performance.

Table~\ref{tab:perturbations} shows the results for the best-performing detector Gragnaniello2021 trained on ProGAN images.
Note that this detection method employs training augmentation using blurring and JPEG compression.
We copy the results on ``clean'' images from Table~\ref{tab:pretrained} as a reference.
For GAN-generated images, blurring and cropping have only a very small effect on the detection performance, while compression and the addition of noise cause a stronger deterioration (\SI{1}{\percent} and \SI{3.4}{\percent} average AUROC decrease, respectively).
Overall, we observe that perturbations (except for cropping) have a stronger effect on DM-generated images compared to GAN-generated images.
On average, the AUROC decreases by \SI{12.66}{\percent} for blurring, \SI{14.98}{\percent} for compression, and \SI{12.18}{\percent} for noise.
We repeat the experiment using the detector Wang2020 fine-tuned on images from all DMs (see \cref{app:finetuning}).
As the results in Table~\ref{tab:perturbations-ft} show, the effect of blurring and cropping is now almost negligible.
The performance drop caused by JPEG compression is also significantly smaller (\SI{2.42}{\percent} average AUROC decrease).
While fine-tuning improves the results for these three perturbations, the same does not hold for adding Gaussian noise.
The latter could be related to the fact that the image generation process of DMs involves noise.
However, another potential explanation is that the detector is shown blurred and compressed images during training but not those with added noise.

\begin{table}[ht]
    \footnotesize
    \centering
    \caption{Effect of image perturbations on detection performance. We use the pre-trained detector Gragnaniello2021 trained on ProGAN images and compute metrics from \num{10000} samples.}
    \label{tab:perturbations}
    \begin{tabular}{lccccc}
    \toprule
    AUROC / Pd@1\% & Clean & Blur & Crop & JPEG & Noise \\ \midrule
    ProGAN            & 100.0 / 100.0                                         & 100.0 / 100.0 & 100.0 / 100.0 & 100.0 / \phantom{0}98.6 & \phantom{0}99.2 / \phantom{0}82.5 \\
    StyleGAN          & 100.0 / 100.0                                         & \phantom{0}99.7 / \phantom{0}94.3 & 100.0 / \phantom{0}99.9 & \phantom{0}99.0 / \phantom{0}79.9 & \phantom{0}94.3 / \phantom{0}40.9 \\
    ProjectedGAN      & 100.0 / \phantom{0}99.3                     & \phantom{0}99.2 / \phantom{0}87.6 & \phantom{0}99.9 / \phantom{0}98.0 & \phantom{0}98.2 / \phantom{0}71.7 & \phantom{0}96.6 / \phantom{0}52.8 \\
    Diff-StyleGAN2    & 100.0 / 100.0                                         & 100.0 / 100.0 & 100.0 / 100.0 & \phantom{0}99.7 / \phantom{0}92.7 & \phantom{0}96.7 / \phantom{0}56.3 \\
    Diff-ProjectedGAN & \phantom{0}99.9 / \phantom{0}99.2           & \phantom{0}99.0 / \phantom{0}84.1 & \phantom{0}99.9 / \phantom{0}97.0 & \phantom{0}98.1 / \phantom{0}70.6 & \phantom{0}96.1 / \phantom{0}51.0 \\ \midrule
    DDPM              & \phantom{0}96.5 / \phantom{0}39.1           & \phantom{0}78.5 / \phantom{0}\phantom{0}8.1 & \phantom{0}95.5 / \phantom{0}35.2 & \phantom{0}81.1 / \phantom{0}\phantom{0}6.3 & \phantom{0}85.1 / \phantom{0}11.2 \\
    IDDPM             & \phantom{0}94.3 / \phantom{0}25.7           & \phantom{0}75.3 / \phantom{0}\phantom{0}5.5 & \phantom{0}93.5 / \phantom{0}24.6 & \phantom{0}77.5 / \phantom{0}\phantom{0}5.0 & \phantom{0}80.9 / \phantom{0}\phantom{0}7.0 \\
    ADM               & \phantom{0}77.8 / \phantom{0}\phantom{0}5.2 & \phantom{0}66.0 / \phantom{0}\phantom{0}3.0 & \phantom{0}78.3 / \phantom{0}\phantom{0}5.5 & \phantom{0}64.1 / \phantom{0}\phantom{0}1.6 & \phantom{0}64.8 / \phantom{0}\phantom{0}1.7 \\
    PNDM              & \phantom{0}91.6 / \phantom{0}16.6           & \phantom{0}86.7 / \phantom{0}14.3 & \phantom{0}91.1 / \phantom{0}19.7 & \phantom{0}76.0 / \phantom{0}\phantom{0}3.9 & \phantom{0}81.2 / \phantom{0}\phantom{0}9.3 \\
    LDM               & \phantom{0}96.7 / \phantom{0}42.1           & \phantom{0}87.1 / \phantom{0}17.1 & \phantom{0}96.8 / \phantom{0}48.9 & \phantom{0}83.3 / \phantom{0}\phantom{0}8.3 & \phantom{0}82.2 / \phantom{0}\phantom{0}8.8 \\ \bottomrule
\end{tabular}

\end{table}

\begin{table}[ht]
    \footnotesize
    \centering
    \caption{Effect of image perturbations on detection performance. We use the detector Wang2020 fine-tuned on all DM-generated images and compute metrics from \num{10000} samples.}
    \label{tab:perturbations-ft}
    \begin{tabular}{lccccc}
    \toprule
    AUROC / Pd@1\%     & Clean                           & Blur                                                & Crop                            & JPEG                          & Noise                                               \\ \midrule
    DDPM  & 100.0 / 100.0           & 100.0 / \phantom{0}99.9                     & 100.0 / \phantom{0}99.9 & \phantom{0}98.5 / \phantom{0}80.1 & \phantom{0}69.8 / \phantom{0}11.3 \\
    IDDPM & 100.0 / 100.0           & 100.0 / 100.0                               & 100.0 / 100.0           & \phantom{0}98.1 / \phantom{0}75.8 & \phantom{0}70.2 / \phantom{0}11.1 \\
    ADM   & 100.0 / \phantom{0}99.7 & \phantom{0}99.9 / \phantom{0}99.0 & 100.0 / \phantom{0}99.5 & \phantom{0}94.5 / \phantom{0}47.1 & \phantom{0}70.3 / \phantom{0}10.8 \\
    PNDM  & 100.0 / 100.0           & 100.0 / \phantom{0}99.8           & 100.0 / 100.0           & \phantom{0}98.7 / \phantom{0}84.5 & \phantom{0}74.8 / \phantom{0}14.6 \\
    LDM   & 100.0 / 100.0           & 100.0 / 100.0                               & 100.0 / 100.0           & \phantom{0}98.1 / \phantom{0}77.3 & \phantom{0}71.9 / \phantom{0}12.9  \\
    \bottomrule
\end{tabular}

\end{table}

\subsection{Fakeness Rating}
\label{app:fakeness}
\begin{figure}[hb]
    \centering
    \subfloat[DDPM]{%
        \centering
        \includegraphics[scale=\imagescale]{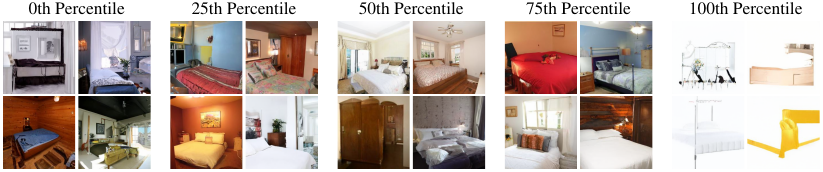}%
    }
    \vspace{2mm}
    \subfloat[IDDPM]{%
        \centering
        \includegraphics[scale=\imagescale]{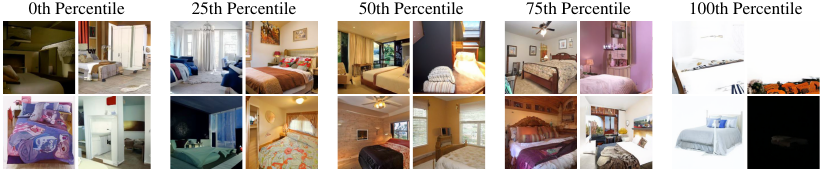}%
    }
    \vspace{2mm}
    \subfloat[ADM]{%
        \centering
        \includegraphics[scale=\imagescale]{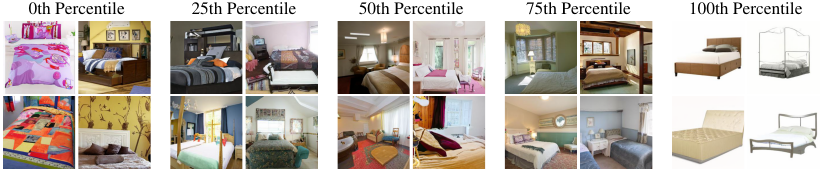}%
    }
    \vspace{2mm}
    \subfloat[PNDM]{%
        \centering
        \includegraphics[scale=\imagescale]{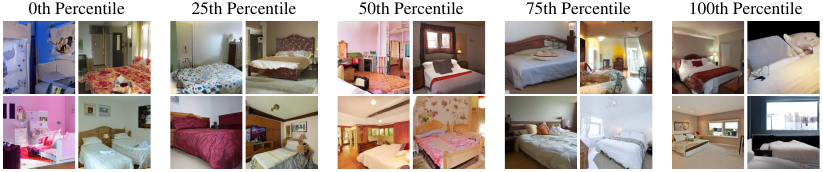}%
    }
    \vspace{2mm}
    \subfloat[LDM]{%
        \centering
        \includegraphics[scale=\imagescale]{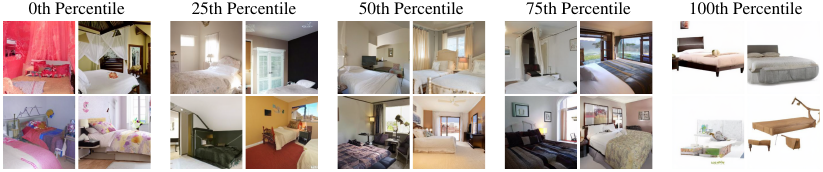}%
    }
    \caption{``Fakeness'' rating based of DM-generated images (LSUN Bedroom) on predictions from detector Gragnaniello2021 trained on ProGAN images. Images are ranked by the model's output, i.e., images in the 100th percentile are considered most fake.}
    \label{fig:fakeness-pretrained}
\end{figure}

\begin{figure}%
    \centering
    \subfloat[DDPM]{
        \centering
        \includegraphics[scale=\imagescale]{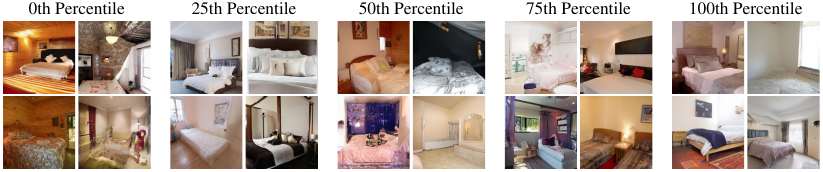}
    }
    \vspace{2mm}
    \subfloat[IDDPM]{
        \centering
        \includegraphics[scale=\imagescale]{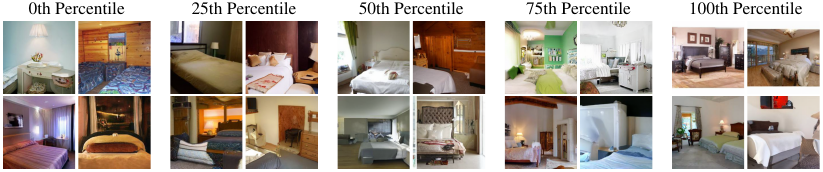}
    }
    \vspace{2mm}
    \subfloat[ADM]{
        \centering
        \includegraphics[scale=\imagescale]{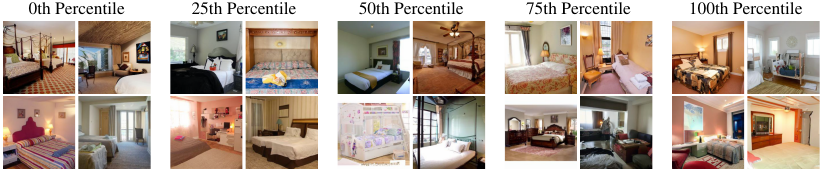}
    }
    \vspace{2mm}
    \subfloat[PNDM]{
        \centering
        \includegraphics[scale=\imagescale]{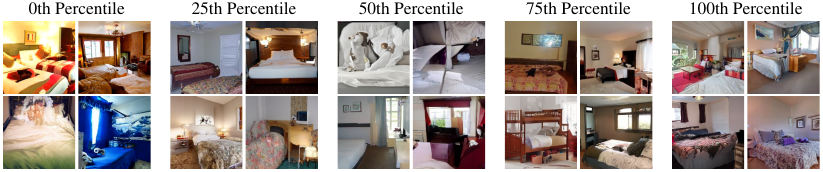}
    }
    \vspace{2mm}
    \subfloat[LDM]{
        \centering
        \includegraphics[scale=\imagescale]{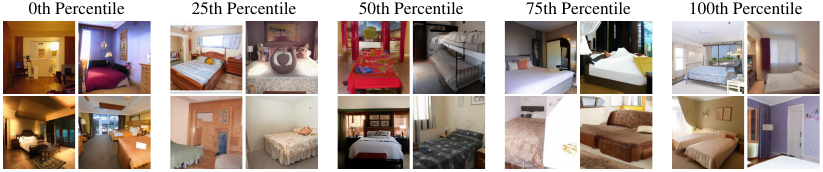}
    }
    \caption{``Fakeness'' rating of DM-generated images (LSUN Bedroom) based on predictions from detector Wang2020 fine-tuned on images from all DMs. Images are ranked by the model's output, i.e., images in the 100th percentile are considered most fake.}
    \label{fig:fakeness-finetuned}
\end{figure}

\begin{figure}%
    \centering
    \subfloat[LDM (FFHQ)]{%
        \centering
        \includegraphics[scale=\imagescale]{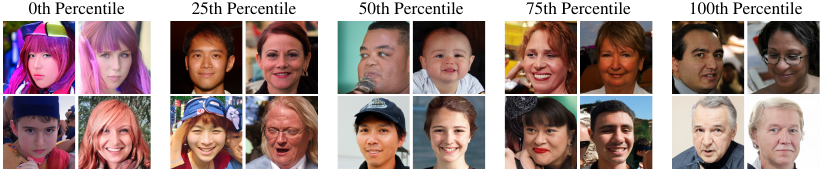}%
    }
    \vspace{2mm}
    \subfloat[ADM' (FFHQ)]{%
        \centering
        \includegraphics[scale=\imagescale]{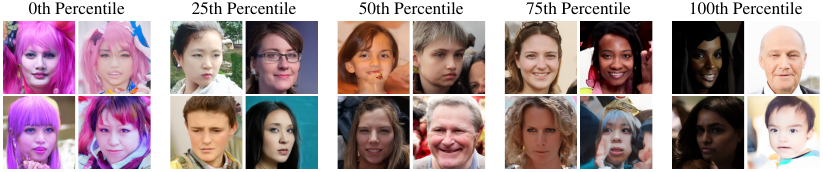}%
    }
    \vspace{2mm}
    \subfloat[ADM (ImageNet)]{%
        \centering
        \includegraphics[scale=\imagescale]{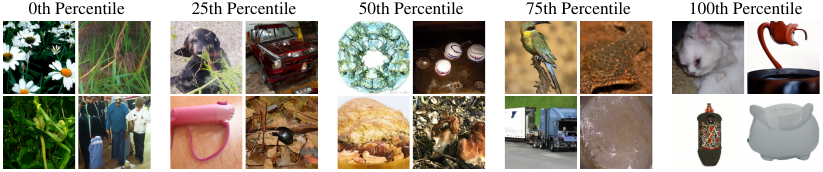}%
    }
    \vspace{2mm}
    \subfloat[PNDM (LSUN Church)]{%
        \centering
        \includegraphics[scale=\imagescale]{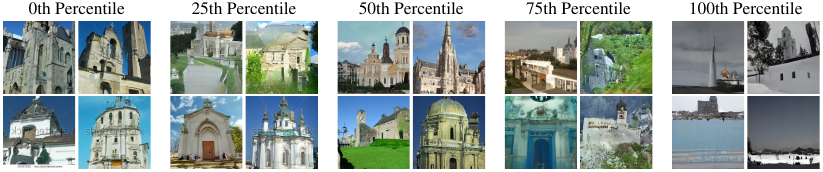}%
    }
    \caption{``Fakeness'' rating of DM-generated images (other datasets) based on predictions from detector Gragnaniello2021 trained on ProGAN images. Images are ranked by the model's output, i.e., images in the 100th percentile are considered most fake.}
    \label{fig:fakeness-other}
\end{figure}

We evaluate whether DM-generated images exhibit some visual cues used by the detector to distinguish them from real images.
Inspired by \cite{wangCNNgeneratedImagesAre2020}, we rank all images by the model's predictions (higher value means ``more fake'') and show examples from different percentiles.
We consider two detectors, the pre-trained detector Gragnaniello2021 trained on ProGAN (Figure~\ref{fig:fakeness-pretrained}) and the detector Wang2020 fine-tuned on all DM-generated images (Figure~\ref{fig:fakeness-finetuned}).
Using Gragnaniello2021, we make the observation that for most DMs, images which the model assigns a high ``fakeness'' score contain many pixels which are purely white or black.
On the other hand, images considered less fake appear to be more colorful.
We were able to reproduce this behavior for other datasets in Figure~\ref{fig:fakeness-other}.
However, this finding does not hold for the ranking provided by the fine-tuned detector. %
Overall, we agree with~\cite{wangCNNgeneratedImagesAre2020} that there is no strong correlation between the model's predictions and visual quality.

\clearpage
\section{\uppercase{Supplemental Material on the Frequency Analysis}}
\label{app:frequency}

\subsection{Descriptions of Frequency Transforms}
\label{app:transforms}
\paragraph{Discrete Fourier Transform (DFT)}
The \ac{DFT} maps a discrete signal to the frequency domain by expressing it as a sum of periodic basis functions.
Given a grayscale image $I$ with height $H$ and width $W$, the two-dimensional \ac{DFT} (with normalization term omitted) is defined as
\begin{equation}
	I_\text{DFT}[k,l] = \sum_{x=0}^{H-1}\sum_{y=0}^{W-1} I[x, y] \; \exp^{-2\pi i \frac{x\cdot k}{H}} \; \exp^{-2\pi i \frac{y\cdot l}{W}},
\end{equation}
with $k=0,\dots,H-1$ and $l=0,\dots,W-1$.
For visualization, the zero-frequency component is shifted to the center of the spectrum.
Therefore, coefficients towards the edges of the spectrum correspond to higher frequencies.

\paragraph{Discrete Cosine Transform (DCT)}
The \ac{DCT} is closely related to the \ac{DFT}, however it uses real-valued cosine functions as basis functions.
It is used in the JPEG compression standard due to its high degree of energy compaction \cite{wallaceJPEGStillPicture1991}, which ensures that a large portion of a signal's energy can be represented using only a few \ac{DCT} coefficients.
The type-II \ac{DCT}, which the term \ac{DCT} usually refers to, is given as
\begin{equation}
    I_\text{DCT}[k, l] = \sum_{x=0}^{H - 1} \sum_{y=0}^{W - 1} I[x, y] \; \text{cos} \bigg[ \frac{\pi}{H} \bigg( x + \frac{1}{2} \bigg)  k_x \bigg] \; \text{cos} \bigg[ \frac{\pi}{W} \bigg(y + \frac{1}{2} \bigg) k_y \bigg],
\end{equation}
again omitting the normalization factor.
In the resulting spectrum, the low frequencies are located in the upper left corner, with frequencies increasing along both spatial dimensions.

\paragraph{Reduced Spectrum}
While the \ac{DFT} provides a useful visual representation of an image's spectrum, it is less suitable for comparing different spectra quantitatively.
Therefore, previous works use the reduced spectrum, a one-dimensional representation of the Fourier spectrum \cite{durallWatchYourUpconvolution2020,dzanicFourierSpectrumDiscrepancies2020,schwarzFrequencyBiasGenerative2021,chandrasegaranCloserLookFourier2021}.
The definition of the reduced spectrum slightly differs between different existing works, we decide to follow that of Schwarz et al.~\cite{schwarzFrequencyBiasGenerative2021}.
It is obtained by azimuthally averaging over the spectrum in normalized polar coordinates $r \in [0, 1]$, $\theta \in [0, 2\pi)$ according to

\begin{equation}
	\tilde{S}(r) = \frac{1}{2\pi}\int_0^{2\pi} S(r,\theta) d\theta \quad \text{with} \quad r = \sqrt{\frac{k^2+l^2}{\frac{1}{4}(H^2 + W^2)}}\quad \text{and} \quad \theta=\mathrm{atan2}(k,l),
\end{equation}
with $S[k, l] = |I_{DFT}[k, l]|^2$ being the squared magnitudes of the Fourier coefficients.
The maximum frequency is given by the Nyquist frequency $f_\text{nyq} = \sqrt{k^2 + l^2} = H / \sqrt{2}$ for a square image with $H = W$.

\subsection{DCT Spectra}
\label{app:dct}

\begin{figure}[ht!]
    \centering
    \subfloat[\acp{GAN}]{%
        \centering
        \includegraphics[scale=\imagescale]{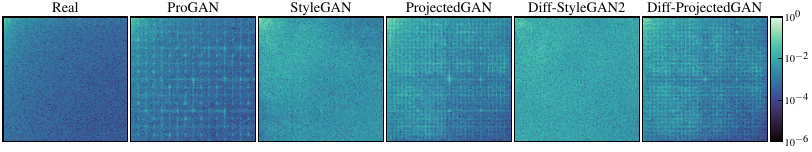}%
    }
    \vspace{2mm}
    \subfloat[\acp{DM}]{%
        \centering
        \includegraphics[scale=\imagescale]{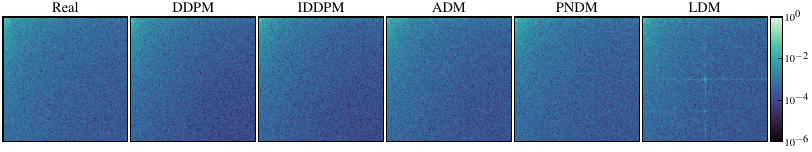}%
    }
    \caption{Mean \ac{DCT} spectrum of real and generated images. To increase visibility, the color bar is limited to $[10^{-6}, 10^{0}]$, with values lying outside this interval being clipped.}
    \label{fig:dct}
\end{figure}

Figure~\ref{fig:dct} depicts the \ac{DCT} spectra of both \acp{GAN} and \acp{DM}.
Similar to \ac{DFT}, images generated by \acp{DM} exhibit fewer artifacts, except for LDM.

\subsection{Logistic Regression Experiments}
\label{app:logistic_regression}
The experiments in Section \ref{sec:frequency}, in particular Figures~\ref{fig:dft} and \ref{fig:reduced}, demonstrate that although \acp{DM} do not exhibit strong frequency artifacts, their spectrum deviates from that of real images.
A natural question is therefore whether these discrepancies can be used to detect \ac{DM}-generated images more effectively.
Following the work of~\cite{frankLeveragingFrequencyAnalysis2020}, we perform a simple logistic regression on each dataset with different transforms: Pixel (no transform), \ac{DFT} (and taking the absolute value), and \ac{DCT}.

We use \num{20000} samples for training, \num{2000} for validation, and \num{20000} for testing, each set equally split between real and fake images.
To reduce the number of features, we transform all images to grayscale and take a center crop with dimensions 64$\times$64.
Additionally, all features are independently standardized to have zero mean and unit variance.
We apply $L_2$ regularization and identify the optimal regularization weight by performing a grid search over the range $\big\{10^{k}\ \mid\ k \in \{-4, -3, \dots, 4\}\big\}$.

\begin{table}%
    \footnotesize
    \centering
    \caption{Accuracy of logistic regression on pixels and different transforms. Next to each transform column we report the gain compared to the accuracy on pixels.}
    \label{tab:logistic_regression}
    \begin{tabular}{lrrrrrrrrr}
    \toprule
                      & Pixel &           \multicolumn{2}{l}{DFT}  & \multicolumn{2}{l}{log(DFT)}  &  \multicolumn{2}{l}{DCT}          &      \multicolumn{2}{l}{log(DCT)}  \\ \midrule
    ProGAN            &  64.8 &          74.7 &  \color{gray} +9.9 &     72.6 &  \color{gray} +7.8 &          65.4 & \color{gray} +0.6 & \textbf{74.9} & \color{gray} +10.1 \\
    StyleGAN          &  91.1 &          87.4 &  \color{gray} -3.7 &     86.2 &  \color{gray} -4.9 & \textbf{92.6} & \color{gray} +1.5 &          86.4 &  \color{gray} -4.7 \\
    ProjectedGAN      &  90.0 &          90.8 &  \color{gray} +0.8 &     90.3 &  \color{gray} +0.3 &          91.0 & \color{gray} +1.0 & \textbf{95.3} &  \color{gray} +5.3 \\
    Diff-StyleGAN2    &  92.4 &          80.3 & \color{gray} -12.0 &     80.5 & \color{gray} -11.9 & \textbf{93.8} & \color{gray} +1.4 &          87.6 &  \color{gray} -4.7 \\
    Diff-ProjectedGAN &  87.4 &          93.9 &  \color{gray} +6.5 &     93.1 &  \color{gray} +5.7 &          88.1 & \color{gray} +0.6 & \textbf{97.7} & \color{gray} +10.3 \\ \midrule
    DDPM              &  51.7 &          64.2 & \color{gray} +12.6 &     64.2 & \color{gray} +12.5 &          52.4 & \color{gray} +0.7 & \textbf{64.3} & \color{gray} +12.6 \\
    IDDPM             &  51.6 & \textbf{62.1} & \color{gray} +10.6 &     61.7 & \color{gray} +10.1 &          51.7 & \color{gray} +0.1 &          61.7 & \color{gray} +10.1 \\
    ADM               &  50.1 & \textbf{54.7} &  \color{gray} +4.6 &     52.3 &  \color{gray} +2.3 &          50.1 & \color{gray} +0.0 &          53.8 &  \color{gray} +3.7 \\
    PNDM              &  52.5 &          57.0 &  \color{gray} +4.5 &     58.2 &  \color{gray} +5.7 &          51.1 & \color{gray} -1.4 & \textbf{61.4} &  \color{gray} +8.9 \\
    LDM               &  56.6 &          63.7 &  \color{gray} +7.1 &     66.1 &  \color{gray} +9.5 &          58.5 & \color{gray} +1.9 & \textbf{73.0} & \color{gray} +16.3 \\ \bottomrule
\end{tabular}

\end{table}

The accuracy of all \acp{GAN} and \acp{DM} in our dataset is given in Table~\ref{tab:logistic_regression}.
We also report the results for log-scaled \ac{DFT} and \ac{DCT} coefficients, as this leads to significant improvements for some generators.
For both GANs and DMs, using information from the frequency domain increases classification accuracy.
On average, the performance gain of the best transform compared to no transform is \SI{5.72}{\percent} and \SI{10.6}{\percent}, respectively.
Although the gain for \acp{DM} is more than double that for \acp{GAN}, the overall maximum accuracy is significantly lower (\SI{90.9}{\percent} for \acp{GAN} and \SI{63.1}{\percent} for \acp{DM} on average).
Therefore, we can not conclude that \acp{DM} exhibit stronger discriminative features in the frequency domain compared to \acp{GAN}.
However, these results corroborate the hypothesis that \ac{DM}-generated images are more difficult to detect.

\subsection{Spectrum Evolution During the Denoising Process} 
\label{app:denoising}
\begin{figure}
    \centering
    \hspace*{\fill}%
    \subfloat[$0 \le t \le 1000$]{%
        \centering
        \includegraphics[scale=\imagescale]{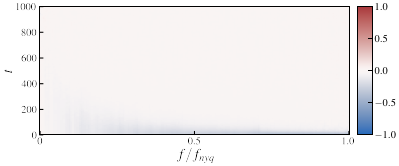}%
    }
    \hspace*{\fill}%
    \subfloat[$0 \le t \le 100$]{%
        \centering
        \includegraphics[scale=\imagescale]{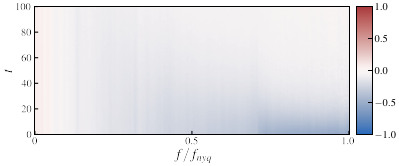}%
    }
    \hspace*{\fill}%
    \caption{Evolution of spectral density error throughout the denoising process. The error is computed relatively to the corresponding step in the diffusion process. The colorbar is clipped at -1 and 1. If the model was able to perfectly denoise, the difference would be 0 at all $t$.}
    \label{fig:denoising_vs_diffusion}
\end{figure}

Analogous to Figure~\ref{fig:denoising}, we show the spectrum error evolution during the denoising process in Figure~\ref{fig:denoising_vs_diffusion}.
Here, however, the error is computed relative to the spectrum of noised images at the corresponding step during the diffusion process.
Similar to the denoising process, the spectra of the diffusion process are averaged over 512 samples.
While the relative error is close to zero for a long time during the denoising process, the model fails to accurately reproduce higher frequencies towards $t=0$.
More precisely, too many high-frequency components are removed.%

\begin{figure}
    \centering
    \includegraphics[scale=\imagescale]{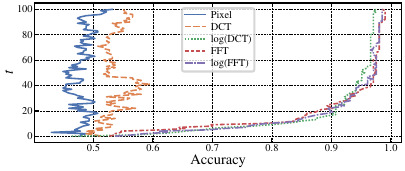}
    \caption{Accuracy of logistic regression during the denoising process. Note that only the last 100 steps of the denoising process are depicted.}
    \label{fig:denoising_lr}
\end{figure}

\begin{figure}
    \centering
    \includegraphics[scale=\imagescale]{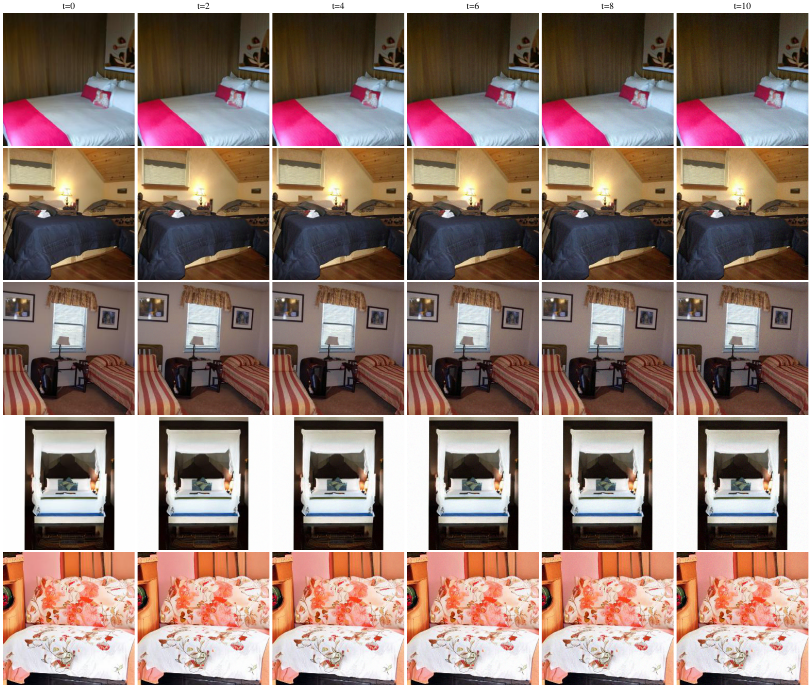}
    \caption{Example images generated by ADM at different $\mathbf{t}$. When zoomed in, the high-frequency noise towards higher $t$ becomes apparent.}
    \label{fig:denoising_examples}
\end{figure}

One could think that, by stopping the denoising process early, \ac{DM}-generated images might be harder to detect.
To test this hypothesis, we perform a logistic regression at every $0 \le t \le 100$ to distinguish real from increasingly denoised generated images.
The model is trained using 512 real and 512 generated samples, from which 20\% are used for testing.
We select the optimal regularization weight by performing a 5-fold cross-validation over $\big\{10^{k}\ \mid\ k \in \{-4, -3, \dots, 4\}\big\}$.
Similar to Section~\ref{app:logistic_regression}, we compare the performance on pixels and different transforms.
The results in Figure~\ref{fig:denoising_lr} do not indicate that around $t=10$ fake images are less detectable.
In Figure~\ref{fig:denoising_examples} it becomes apparent that at this $t$ the images are noticeably noisier, which probably explains the increased accuracy.

\subsection{Extended Analysis of Spectrum Discrepancies}
\label{app:freq_source}
Our analysis in the frequency domain (see Figure~\ref{fig:reduced} in the main paper) suggests that current state-of-the-art \acp{DM} do not match the high-frequency content well.
To further analyze these findings, we build on insights from the denoising autoencoder (DAE) literature.
Roughly speaking, the task DAEs face is conceptually similar to the task of the noise predictor $\epsilon_\theta$ in \acp{DM} at a single time step: denoising a disturbed input at a fixed noise level.
Note that while we believe that \acp{DM} and DAEs can be conceptually related, the concepts are distinct in several ways: \acp{DM} use parameter sharing to perform noise prediction at multiple noise levels to set up the generation as an iterative process.
On the other hand, DAEs make use of a latent space to learn suitable representations for reconstruction, classically at a fixed noise level.
Nevertheless, we are convinced that it may be useful to take these insights into account.

A handy observation from DAEs relates the level of corruption to the learned feature scales:
Denoising an image at small noise levels %
requires to accurately model fine granular details of the image, while coarse/large-scale features can still be recovered at high noise levels (see e.g., \cite{vincentStackedDenoisingAutoencoders2010,gerasScheduledDenoisingAutoencoders2015,gerasCompositeDenoisingAutoencoders2016}).
Transferring this insight to DMs, we observe that the training objective guides the reconstruction performance across the different noise levels based on a weighting scheme $w(t)$ that translates to the (relative) importance of certain noise levels throughout the optimization process.

\begin{figure}[h!]
    \centering
    \hspace*{\fill}%
    \subfloat[$0 \le t \le 1000$]{%
        \centering
        \includegraphics[scale=\imagescale]{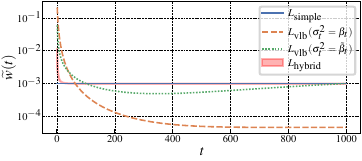}%
    }
    \hspace*{\fill}%
    \subfloat[$0 \le t \le 50$]{%
        \centering
        \includegraphics[scale=\imagescale]{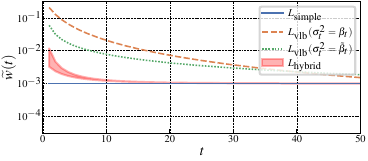}%
    }
    \hspace*{\fill}%
    \caption{Relative importance of the reconstruction tasks for prominent DM loss functions.
    We show the relative influence $\widetilde w(t) = w(t) / \sum_{t=0}^T w(t)$ in Equation \ref{eq:objective} for (a) the whole denoising-diffusion process and (b) a close-up on the lowest noise levels.
    For $L_\mathrm{vlb}$ we depict both, lower and upper bound for the denoising variance $\sigma_t^2$ (given as $\sigma_t^2 = \beta_t$ and $\sigma_t^2 = \tilde \beta_t$ as defined by Ho~et~al.~\protect\cite{hoDenoisingDiffusionProbabilistic2020}).
    In the case of $L_\mathrm{hybrid}$, used by IDDPM and ADM, we use $\sigma_t^2 \mathbf{I}$ for visualization purposes instead of a diagonal covariance matrix and plot $\widetilde w(t)$ for the range of admissible variances $\sigma^2_t$.
    All plots assume the linear noise schedule $\beta_t$ by Ho~et~al.~\protect\cite{hoDenoisingDiffusionProbabilistic2020} and $T=1000$.}
    \label{fig:relative_importance}
\end{figure}

Recall that the objective of many prominent DMs can be stated as a weighted sum of mean squared error (MSE) terms
\begin{equation}
\label{eq:objective}
    L(\theta) = \sum_{t=0}^{T} w(t) \E_{t,\rvx_0,\epsilon} [\Vert \epsilon - \epsilon_\theta(\rvx_t, t) \Vert^2]
\end{equation}
usually with $T=1000$.
The theoretically derived variational lower bound $L_\mathrm{vlb}$, which would optimize for high likelihood values, corresponds to the weighting scheme (here in the case of $\Sigma(t) = \sigma^2_t \mathbf{I}$)
\begin{equation}
    w(t) = \frac{\beta_t^2}{2 \sigma_t^2 \alpha_t ( 1-\bar\alpha_t)}
\end{equation}
with $\alpha_t$ and $\bar\alpha_t$ derived from the noise schedule $\beta_t$ (see Section \ref{sec:background}). 
However, $L_\mathrm{vlb}$ turns out to be extremely difficult to optimize in practice (see e.g., \cite{dhariwalDiffusionModelsBeat2021}), arguably due to large influence of the challenging denoising steps near $t=0$. To circumvent this issue, $L_\mathrm{simple}$ was proposed by Ho~et~al.~\cite{hoDenoisingDiffusionProbabilistic2020} which corresponds to $w(t) = 1$, and turned out to be stable to optimize and already leads to remarkable perceptual quality.
In order to achieve both, perceptual image quality and high likelihood values, Nichol~et~al.~\cite{nicholImprovedDenoisingDiffusion2021} proposed $L_\mathrm{hybrid} = L_\mathrm{simple} + \lambda L_\mathrm{vlb}$ with $\lambda = 0.001$ which increases the influence of low noise levels.
The relative importance of reconstruction tasks at different noise levels for the above discussed objectives are depicted in Figure \ref{fig:relative_importance}.

To put the above discussion in a nutshell, we hypothesize that the ability of a \ac{DM} to match the real frequency spectrum is governed by the reconstruction performance at the corresponding noise levels.
Importantly, successful error prediction at low noise scales, i.e., near $t=0$, requires capturing the high-frequency content of an image (which would be the case when (successfully) optimizing $L_\mathrm{vlb}$).
We deduce that the relative down-weighting of the influence of low noise levels when using $L_\mathrm{simple}$ or $L_\mathrm{hybrid}$ (compared to the theoretically derived $L_\mathrm{vlb}$) results in the observed mismatch of high frequencies.

The mean reduced spectra of various DMs (Figure~\ref{fig:reduced}) support this hypothesis: Both IDDPM and ADM are trained with $L_\mathrm{hybrid}$ (which incorporates a relatively higher weight on low noise levels), and are able to reduce the gap to the real spectrum when compared to the baseline (DDPM) trained with $L_\mathrm{simple}$ \cite{hoDenoisingDiffusionProbabilistic2020}. Clearly, we believe that not only the weighting scheme $w(t)$ accounts for the resulting spectral properties, but more importantly the model's capabilities to successfully predict the low level noise in the first place. From this perspective, the weighting scheme acts as a proxy that encourages the model to focus on specific noise levels.

We conclude that the objectives of \acp{DM} are well-designed to guide the model to high perceptual quality (or benchmark metrics such as FID), while falling short on providing sufficient information to accurately model the high frequency content of the target images, which would be better captured by a likelihood-based objective like $L_\mathrm{vlb}$.

\subsection{Effect of Number of Sampling Steps}
\label{app:steps}
\begin{figure}[h]
    \centering
    \hspace*{\fill}%
    \subfloat[Without DDIM]{%
        \centering
        \includegraphics[scale=\imagescale]{results/sampling_analysis/spectrum_evolution-default-across.pdf}%
    }
    \hspace*{\fill}%
    \subfloat[With DDIM]{%
        \centering
        \includegraphics[scale=\imagescale]{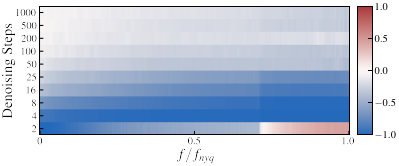}%
    }
    \hspace*{\fill}%
    \caption{Spectral density error $\tilde{S}_\text{err}$ for different numbers of denoising steps. The error is computed relatively to the spectrum of real images. The colorbar is clipped at -1 and 1. Note that the $y$-axis is not scaled linearly.}
    \label{fig:sampling_steps}
\end{figure}

Analogous to \cref{fig:evolution_steps} we show the spectral density error using DDIM~\cite{songDenoisingDiffusionImplicit2022} in \cref{fig:sampling_steps}.
As a reference, we also include the error without DDIM.
Performing more steps leads to lower errors, with DDIM improving faster than normal sampling.
This finding is coherent with previous results, as DDIM generates better samples at fewer sampling steps \cite{songDenoisingDiffusionImplicit2022}.

\begin{figure}[h]
    \centering
    \hspace*{\fill}%
    \subfloat[Without DDIM]{%
        \centering
        \includegraphics[scale=\imagescale]{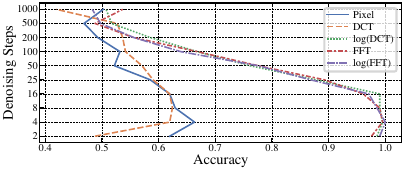}%
    }
    \hspace*{\fill}%
    \subfloat[With DDIM]{%
        \centering
        \includegraphics[scale=\imagescale]{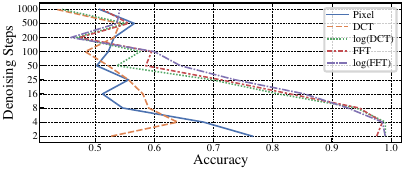}%
    }
    \hspace*{\fill}%
    \caption{Accuracy of logistic regression for different numbers of sampling steps. Note that the y-axis is not scaled linearly.}
    \label{fig:steps_lr}
\end{figure}

\begin{figure}[p]
    \centering
    \subfloat[Without DDIM]{%
        \centering
        \includegraphics[scale=\imagescale]{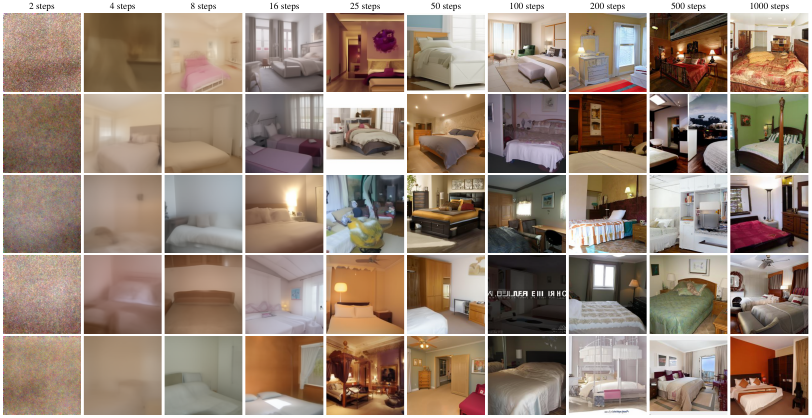}%
    }
    \vspace{2mm}
    \subfloat[With DDIM]{%
        \centering
        \includegraphics[scale=\imagescale]{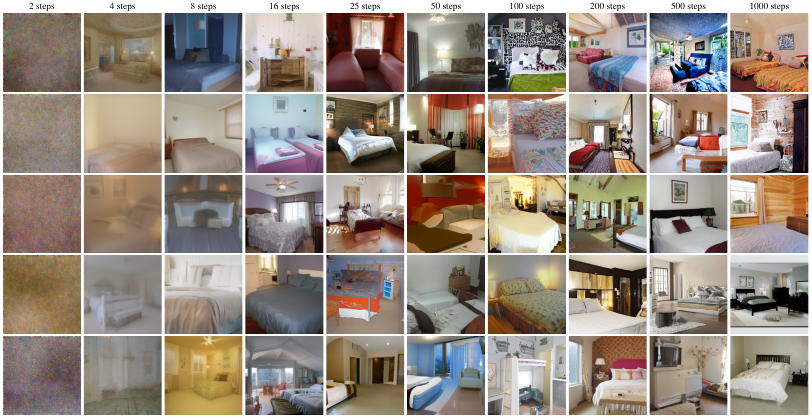}%
    }
    \caption{Example images generated by ADM with different numbers of sampling steps.}
    \label{fig:steps_examples}
\end{figure}

In addition, we repeat the logistic regression experiment from \cref{app:denoising} for samples generated with different numbers of timesteps, the results are shown in Figure~\ref{fig:steps_lr}.
We also provide example images in Figure~\ref{fig:steps_examples}.
As expected, images become harder to detect with more sampling steps and therefore higher image quality.

\clearpage

\subsection{Frequency Analysis on Additional Datasets}
\label{app:additional_frequency}
\begin{figure}[hb!]
    \centering
    \hspace*{\fill}%
    \subfloat[LSUN Cat]{%
        \centering
        \includegraphics[scale=\imagescale]{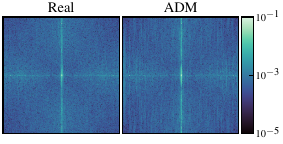}%
    }
    \hspace*{\fill}%
    \subfloat[LSUN Horse]{%
        \centering
        \includegraphics[scale=\imagescale]{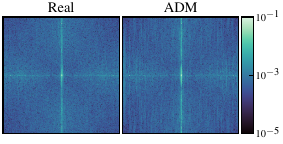}%
    }
    \hspace*{\fill}%
    \vspace{2mm}
    \hspace*{\fill}%
    \subfloat[ImageNet]{%
        \centering
        \includegraphics[scale=\imagescale]{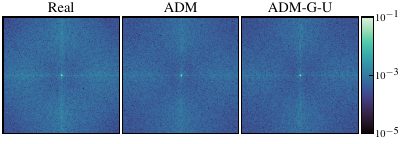}%
    }
    \hspace*{\fill}%
    \subfloat[LSUN Church]{%
        \centering
        \includegraphics[scale=\imagescale]{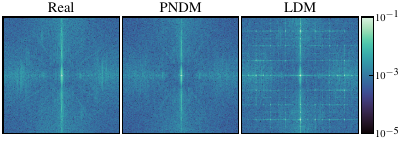}%
    }
    \hspace*{\fill}%
    \vspace{2mm}
    \subfloat[FFHQ (*: pre-processing according to LDM~\protect\cite{rombachHighresolutionImageSynthesis2022}, **: pre-processing according to P2~\protect\cite{choiPerceptionPrioritizedTraining2022})]{%
        \centering
        \includegraphics[scale=\imagescale]{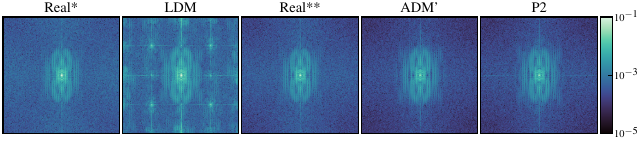}%
    }
    \vspace{2mm}
    \subfloat[Stable Diffusion]{%
        \centering
        \includegraphics[scale=\imagescale]{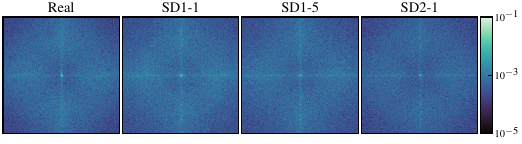}%
    }
    \caption{Mean \ac{DFT} spectrum of real and generated images from additional datasets. To increase visibility, the color bar is limited to $[10^{-5}, 10^{-1}]$, with values lying outside this interval being clipped.}
    \label{fig:dft_other}
\end{figure}

\begin{figure}[h!]
    \centering
    \hspace*{\fill}%
    \subfloat[LSUN Cat]{%
        \centering
        \includegraphics[scale=\imagescale]{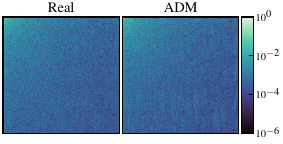}%
    }
    \hspace*{\fill}%
    \subfloat[LSUN Horse]{%
        \centering
        \includegraphics[scale=\imagescale]{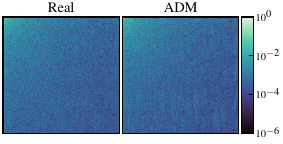}%
    }
    \hspace*{\fill}%
    \vspace{2mm}
    \hspace*{\fill}%
    \subfloat[ImageNet]{%
        \centering
        \includegraphics[scale=\imagescale]{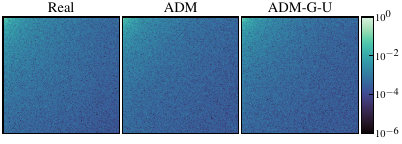}%
    }
    \hspace*{\fill}%
    \subfloat[LSUN Church]{%
        \centering
        \includegraphics[scale=\imagescale]{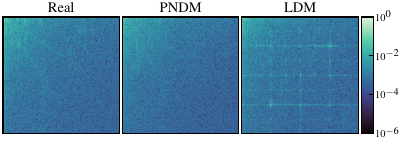}%
    }
    \hspace*{\fill}%
    \vspace{2mm}
    \subfloat[FFHQ (*: pre-processing according to LDM~\protect\cite{rombachHighresolutionImageSynthesis2022}, **: pre-processing according to P2~\protect\cite{choiPerceptionPrioritizedTraining2022})]{%
        \centering
        \includegraphics[scale=\imagescale]{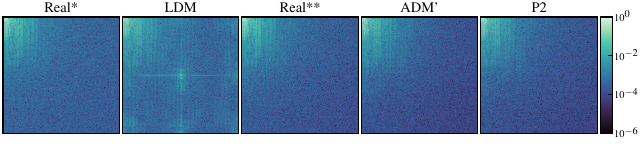}%
    }
    \vspace{2mm}
    \subfloat[Stable Diffusion]{%
        \centering
        \includegraphics[scale=\imagescale]{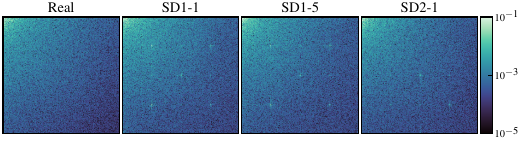}%
    }
    \caption{Mean \ac{DCT} spectrum of real and generated images from additional datasets. To increase visibility, the color bar is limited to $[10^{-6}, 10^{0}]$, with values lying outside this interval being clipped.}
    \label{fig:dct_other}
\end{figure}

\begin{figure}[h!]
    \centering
    \hspace*{\fill}%
    \subfloat[LSUN Cat]{%
        \centering
        \includegraphics[scale=\imagescale]{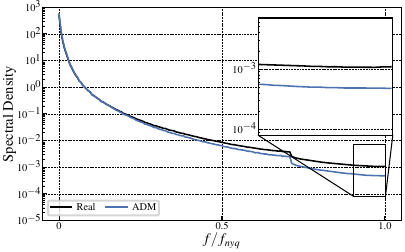}%
    }
    \hspace*{\fill}%
    \subfloat[LSUN Horse]{%
        \centering
        \includegraphics[scale=\imagescale]{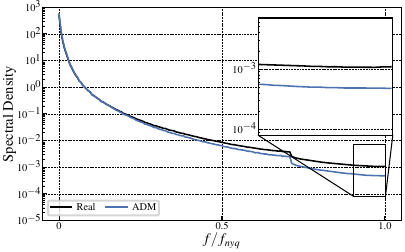}%
    }
    \hspace*{\fill}%
    \vspace{2mm}
    \hspace*{\fill}%
    \subfloat[ImageNet]{%
        \centering
        \includegraphics[scale=\imagescale]{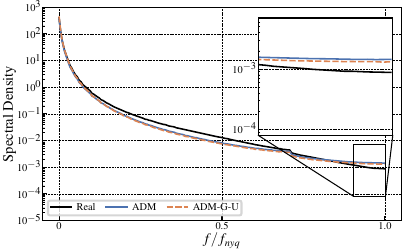}%
    }
    \hspace*{\fill}%
    \subfloat[LSUN Church]{%
        \centering
        \includegraphics[scale=\imagescale]{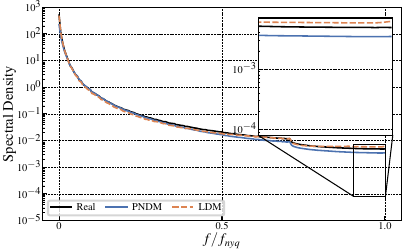}%
    }
    \hspace*{\fill}%
    \vspace{2mm}
    \hspace*{\fill}%
    \subfloat[FFHQ (*: pre-processing according to LDM~\protect\cite{rombachHighresolutionImageSynthesis2022}, **: pre-processing according to P2\protect\cite{choiPerceptionPrioritizedTraining2022})]{%
        \centering
        \includegraphics[scale=\imagescale]{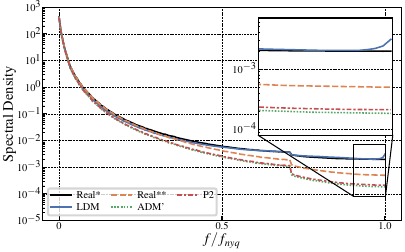}%
    }
    \hspace*{\fill}%
    \subfloat[Stable Diffusion]{%
        \centering
        \includegraphics[scale=\imagescale]{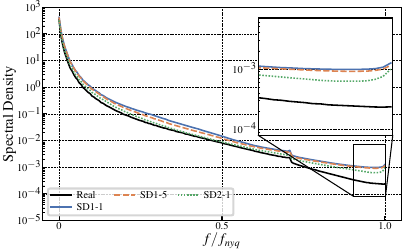}%
    }
    \hspace*{\fill}%
    \caption{Mean reduced spectrum of real and generated images from additional datasets. The part of the spectrum where \ac{GAN}-characteristic discrepancies occur is magnified.}
    \label{fig:reduced_other}
\end{figure}

We analyze the \ac{DFT} (Figure~\ref{fig:dft_other}), \ac{DCT} (Figure~\ref{fig:dct_other}), and reduced spectra (Figure~\ref{fig:reduced_other}) using a similar process as in Section~\ref{sec:properties}.
Regarding frequency artifacts, the results are consistent with that from LSUN Bedroom, LDM exhibits grid-like artifacts while ADM and PNDM do not.
The spectra of ADM on LSUN Cat and LSUN Horse do contain irregular, vertical structures, which we did not observe for any other model and dataset.
However, these are substantially different and not as pronounced as \ac{GAN} artifacts.
The \ac{DFT} and \ac{DCT} of (real and generated) FFHQ images clearly deviate from the remaining spectra, which we attribute to the homogeneity of the dataset.
Note that LDM and ADM'/P2 were trained on differently processed versions of the real images, which is why we include the spectra of both variants.
While we observe the known artifacts for LDM, ADM' and P2 do not contain such patterns.

The reduced spectra have largely the same characteristics as for LSUN Bedroom, except for ImageNet.
Here we observe an overestimation towards higher frequencies, which is the opposite of what we see for ADM on other datasets.
A possible explanation could be that the authors sampled images from LSUN using 1000 and from ImageNet using only 250 steps.
We suppose that the amount of spectral discrepancies is highly training dependent.

As discussed in Section~\ref{sec:frequency}, the \ac{DFT} and \ac{DCT} spectra of images generated by Stable Diffusion exhibit very subtle grid-like artifacts.
Note that we exclude images generated by Midjourney from this analysis due to the small number of available samples.
The reduced spectra show similarities to those of LDM, with a rise towards the higher end of the spectrum.
However, the spectral density of images generated by Stable Diffusion is higher than that of real images throughout the spectrum.
It should be noted that these deviations might be caused due to the different data distributions of real and generated images.

\end{document}